\documentclass[letterpaper,twocolumn,10pt]{article}
\usepackage{usenix}
\usepackage{booktabs}

\usepackage{graphicx}
\usepackage{float}
\usepackage{wrapfig}
\usepackage[normalem]{ulem}

\usepackage{bm}
\usepackage{textcomp}
\usepackage{amsmath}
\usepackage{amsthm}
\usepackage{amsfonts}
\usepackage{mathtools}
\usepackage{pifont}

\usepackage{algorithm}
\usepackage{algorithmic}

\usepackage{times}
\usepackage{gensymb}

\usepackage{textcomp}
\usepackage{multirow}
\usepackage{lscape}
\usepackage{subcaption}

\usepackage{mdframed}
\usepackage{hyperref}
\usepackage{textgreek}
\usepackage{tabularx, colortbl, xcolor}
\definecolor{mygray}{gray}{0.95}
\definecolor{mycyan}{HTML}{005397}
\definecolor{myred}{HTML}{E13333}
\definecolor{mymagenta}{HTML}{BF3E87}
\definecolor{mypurple}{HTML}{1B2278}
\definecolor{tearose}{HTML}{F584C5}

\definecolor{coral}{HTML}{F67088}
\definecolor{dodger_blue}{HTML}{3BA3EC}

\definecolor{domino}{HTML}{BC9F48}
\definecolor{catalina_blue}{HTML}{1C3168}
\definecolor{dark_scarlet}{HTML}{C63D52}
\definecolor{cerulean}{HTML}{0192A8}
\definecolor{tussock}{HTML}{C99E31}
\definecolor{p13}{HTML}{BFB5D7}
\definecolor{b14}{HTML}{BEA1A5}
\definecolor{y15}{HTML}{F0Cf61}
\definecolor{Merino}{HTML}{F3EEE3}

\newcolumntype{a}{>{\columncolor{p13}}l}

\usepackage[capitalize,noabbrev]{cleveref}
\crefname{ineq}{Inequality}{Inequalities}
\creflabelformat{ineq}{#2{\upshape(#1)}#3} 
\crefname{obj}{Objective}{Objectives}
\creflabelformat{obj}{#2{\upshape(#1)}#3} 
\crefname{exp}{Expression}{Expressions}
\creflabelformat{exp}{#2{\upshape(#1)}#3} 

\theoremstyle{remark}

\theoremstyle{plain}
\newtheorem{theorem}{Theorem}[section]
\newtheorem{proposition}[theorem]{Proposition}

\theoremstyle{definition}

\theoremstyle{remark}

\usepackage{braket}
\newcommand{\norm}[1]{\left\lVert#1\right\rVert}
\newcommand{\abs}[1]{\left\lvert#1\right\rvert}

\usepackage{stmaryrd}

\DeclareMathOperator*{\argmax}{arg\,max}
\DeclareMathOperator*{\argmin}{arg\,min}

\usepackage{tikz}
\usetikzlibrary{trees}
\usepackage{tikz-dependency}
\usepackage{tikz-3dplot}
\usetikzlibrary{chains,scopes}
\usetikzlibrary{arrows.meta,automata,positioning}
\usetikzlibrary{shapes.geometric, arrows}
\usepackage{pgfplots}

\pgfplotsset{
  every axis/.append style = {thick},
  tick style = {thick,black},
  /tikz/normal shift/.code 2 args = {%
    \pgftransformshift{%
        \pgfpointscale{#2}{\pgfplotspointouternormalvectorofticklabelaxis{#1}}%
    }%
  },%
  shift/.style = {
    tick align        = outside,
    scaled ticks      = false,
    enlargelimits     = false,
    ticklabel shift   = {#1},
    axis lines*       = left,
    xtick style       = {normal shift={x}{#1}},
    ytick style       = {normal shift={y}{#1}},
    x axis line style = {normal shift={x}{#1}},
    y axis line style = {normal shift={y}{#1}},
  },
  shift/.default = 10pt,
  shift3d/.style = {
    shift=#1,
    ztick style       = {normal shift={z}{#1}},
    z axis line style = {normal shift={z}{#1}},
  },
  shift3d/.default = 10pt,
}

\usepackage{listings}
\usepackage{array}
\newcolumntype{H}{>{\setbox0=\hbox\bgroup}c<{\egroup}@{}}

\begin{document}

\date{}

\title{\Large \bf Towards Certified Probabilistic Robustness with High Accuracy}

\author{
{\rm Ruihan Zhang}\\
Singapore Management University
\and
{\rm Peixin Zhang}\\
Singapore Management University
\and
{\rm Jun Sun}\\
Singapore Management University
} 

\maketitle

\begin{abstract}
Adversarial examples pose a security threat to many critical systems built on neural networks (such as face recognition systems, and self-driving cars). While many methods have been proposed to build robust models, how to build certifiably robust yet accurate neural network models remains an open problem. For example, adversarial training improves empirical robustness, but they do not provide certification of the model's robustness. On the other hand, certified training provides certified robustness but at the cost of a significant accuracy drop. In this work, we propose a novel approach that aims to achieve both high accuracy and certified probabilistic robustness. Our method has two parts, i.e., a probabilistic robust training method with an additional goal of minimizing variance in terms of divergence and a runtime inference method for certified probabilistic robustness of the prediction. The latter enables efficient certification of the model's probabilistic robustness at runtime with statistical guarantees. This is supported by our training objective, which minimizes the variance of the model's predictions in a given vicinity, derived from a general definition of model robustness. Our approach works for a variety of perturbations and is reasonably efficient. Our experiments on multiple models trained on different datasets demonstrate that our approach significantly outperforms existing approaches in terms of both certification rate and accuracy.
\end{abstract}

\section{Introduction}
Neural networks have achieved remarkable success in various applications, including many security-critical systems such as self-driving cars~\cite{kurakin2018adversarial}, and face-recognition-based authentication systems~\cite{sharif2016accessorize}. Unfortunately, several security issues of neural networks have been discovered as well. Arguably the most notable one is the presence of adversarial examples. Adversarial examples are inputs that are carefully crafted by adding human imperceptible perturbation to normal inputs to trigger wrong predictions~\cite{kurakin2016adversarial}. Their existence poses a significant threat when the neural networks are deployed in security-critical scenarios. For example, adversarial examples can mislead road sign recognition systems of self-driving cars and cause accidents~\cite {kurakin2018adversarial}. In other use cases, adversarial examples may allow unauthorized access in face-recognition-based authentication~\cite{sharif2016accessorize}. The increasing adoption of machine learning in security-sensitive domains raises concerns about the robustness of these models against adversarial examples~\cite{papernot2016transferability}.

To tackle the issue of adversarial examples, researchers have developed various mitigation methods. Two well-known categories are adversarial training~\cite{bai2021recent,wong2020fast} and certified training~\cite{muller2022certified,shi2021fast}, both of which aim to improve the robustness of neural networks, i.e., improving their prediction accuracy in the presence of adversarial examples whilst maintaining their accuracy with normal inputs if possible.

Adversarial training works by training the neural network with a mixture of normal and adversarial examples. The latter may be either generated before hand~\cite{miyato2018virtual} or during the training (e.g., min-max training~\cite{zhang2019theoretically}). While empirical studies show that adversarial training often improves model robustness whilst maintaining model accuracy, it does not provide any guarantee of the model robustness~\cite{zhang2019limitations}, which makes them less than ideal. For instance, it has been shown that a model trained through adversarial training remains vulnerable to new threats such as adaptive adversarial attacks~\cite{liu2019adaptiveface,tramer2020adaptive}.

Certified training methods aim to provide a certain guarantee of the robustness of the neural network. These methods typically incorporate robustness verification techniques~\cite{xu2020automatic} during training, i.e., they aim to find a valuation of network parameters such that the model is provably robust with respect to the training samples. While they may be able to certify the model robustness on some input samples, they often reduce the model's accuracy significantly~\cite{chiang2020certified}. Recent studies have shown that state-of-the-art certified defences can result in up to 70\% accuracy drop on MNIST and 90\% on CIFAR-10~\cite{chiang2020certified}.  This is unacceptable for many real-world applications. Therefore, there is a pressing need for a more effective and efficient approach that can achieve both high accuracy and certified robustness. An alternative to certified training is randomized smoothing~\cite{cohen2019certified} which certifies certain form of robustness (e.g., against adversarial attacks within the L$^2$-norm) by systematically introducing noises during training. It however suffers from the same problem of significant accuracy loss.

In this work, we introduce a method that certifies a model's probabilistic robustness whilst maintaining its accuracy. Our method is designed based on the belief that deterministic robustness (i.e., a model is 100\% robust within a certain region) is often infeasible without seriously compromising accuracy, whereas probabilistic robustness (e.g.,  a model makes the same prediction 99\% of the time within a certain region) is often sufficient in practice. Our approach comprises two parts: a probabilistic robust training method that minimizes divergence variance and a runtime inference method to certify the model's robustness. In the training phase, our approach focuses on minimizing variance across model predictions on similar inputs to improve the robustness. Unlike other methods that focus on one specific group of adversarial attacks, e.g., PGD-based adversarial training~\cite{zhang2019theoretically} relies on the PGD attack~\cite{madry2017towards}, our method improves the model's robustness without overfitting to specific adversarial attacks. Furthermore, our method can be easily applied to handle a variety of different perturbations, e.g., such as rotation and scaling. In the inference phase, our approach certifies the model's probabilistic robustness by considering a given input in its peripheral region. We prove that the probabilistic certified robustness of a model can be derived from the accuracy of the model in the peripheral region.

We evaluate our method by training models on multiple standard benchmark datasets and compare them with state-of-the-art robustness-improving methods, including adversarial training, certified training and others. We compare our approach with 13 baseline approaches in terms of standard accuracy (i.e., accuracy on normal test data), adversarial accuracy (i.e., accuracy in the presence of adversarial attacks), certified robustness rate (i.e., the probability of a test sample on which the model's probabilistic robustness is successfully certified), and certified robust accuracy (i.e., probability of a test sample being certified robust and correct). Compared to the state-of-the-art adversarial training, we show that our method achieves a competitive or higher adversarial accuracy while sacrificing significantly less standard accuracy (i.e., up to 50\% less). More importantly, we are able to certify the model robustness with regards to most of the test inputs (i.e., up to 96.8\% on MNIST and 92\% on CIFAR-10). Compared to the state-of-the-art certified training, our method achieves a highly robust model whlist maintaining the model's accuracy, i.e., its standard accuracy is almost twice as high as that of certified training. Overall, the experiments show our method achieves a high level of certified robustness whilst maintaining the model accuracy.

In summary, our contributions include the following. 
\begin{itemize}
    \item A novel training algorithm that improves robustness whilst maintaining high accuracy.
    \item An inference method which works with the above-mentioned training method to provide probabilistic certified robustness.
    \item An extensive evaluation that shows that our method outperforms state-of-the-art adversarial training in terms of robustness and achieves a higher certification rate than all existing methods (while maintaining accuracy within 1\% of that of the normal training).
\end{itemize}

\section{Background and Problem Definition}

In the following, we first briefly introduce relevant concepts, and then review existing methods for enhancing model robustness against adversarial attacks, including adversarial and certified training. Lastly, we define our research problem.

\subsection{Preliminary}

\paragraph{Neural Network Models}
In standard supervised learning, a neural network model is a function that takes inputs from $\mathcal{X}$ and produces outputs in $\mathcal{Y}$, where $\mathcal{X}, \mathcal{Y}$ are sets of inputs and outputs, respectively. Suppose we have a hypothetical function $\bar{h}: \mathcal{X} \to \mathcal{Y}$ that we want to approximate using a neural network model given as $h: \mathcal{X} \to \mathcal{Y}$. For any input $x$ in $\mathcal{X}$, the neural network model $h$ produces a prediction $h(x)$.

With ground-truth label $\bar{h}(x)$, we can compare the deviation of $h(x)$ from $\bar{h}(x)$ using a loss function $\ell(h, x, \bar{h}(x))$. The choice of the loss function depends on the specific problem and data, but common options include the cross-entropy loss for classification and the mean squared error loss for regression. In this work, we focus on neural classification models and leave other models (e.g., generative models) to future work. Hereafter, we write $G_x$ to denote the ground truth for any $x\in\mathcal{X}$.

\paragraph{Criterion of Correct Classification}
Here, inputs are represented as normalized vectors in $\mathcal{X}$. Confidence scores for all $C$ classes are denoted by $\mathcal{Y}$. Normalizing the confidence scores (e.g., using the softmax function) provides the probability of each class, with the predicted class being the one with the highest probability. The loss function $\ell(h, x, G_x)$ measures deviation from the given class $G_x$, and a prediction is \emph{incorrect} if and only if the following is satisfied
\begin{equation}
    \exists~  c\in \set{1, 2, \ldots, C}.\quad c \neq G_x \land \ell(h, x, c) < \ell(h, x, G_x)
\end{equation}

\paragraph{Adversarial Examples}
Adversarial examples pose a security threat to machine learning systems, as they can be maliciously crafted to exploit vulnerabilities in the learned models~\cite{szegedy2013intriguing}. These perturbed inputs are often imperceptible to the human eye but can lead to incorrect predictions or classifications, thus compromising the reliability of the system~\cite{carlini2019evaluating}. The consequences of adversarial attacks can be severe, particularly in security-critical areas. As machine learning models are increasingly integrated into critical applications, such as autonomous vehicles, the risk of adversarial attacks has heightened, with potential consequences ranging from privacy breaches to catastrophic failures~\cite{papernot2016limitations, sharif2016accessorize}.
For instance, in the realm of autonomous vehicles, adversarial perturbations could cause misinterpretation of traffic signs or other crucial elements in the environment, leading to accidents and endangering lives~\cite{evtimov2018robust}.

The existence of an adversarial example can be defined as the presence of two inputs that are nearly identical, but are assigned different classifications by the model. Formally, an adversarial example exists if and only if the following is satisfied.
\begin{equation}
\label[exp]{exp:ae}
    \exists~  x_1, x_2\in\mathcal{X}.~~ d(x_1, x_2)\le \epsilon ~\land~ \argmax h(x_1) \neq \argmax h(x_2) 
\end{equation}
where $d(x_1, x_2)$ denotes a distance measure between the two inputs, and this distance needs to be smaller than a threshold $\epsilon$ to be considered imperceptibly different. Note that the distance function can be defined in a variety of different ways (e.g., the Euclidean distance or the degree of rotation).

\paragraph{Robustness}
The robustness of a neural network model refers to its ability to maintain its prediction in the presence of small perturbations. Formally, if the input data $x$ follows certain distribution $\mathcal{D}$, then $\rho(h)$, the robustness of a model $h$, is defined by the extent to which it maintains predictions in the presence of perturbations to the input data, as quantified by the following formula.
\begin{equation}
\label{eq:robustness}
        P_{x_1\sim\mathcal{D}} \bigg(P_{x_2\sim\mathcal{U}(B(x_1))}\Big(\argmax h(x_1) \neq \argmax h(x_2) \Big) \leq \kappa \bigg)
\end{equation}
where the vicinity~\cite{ma2018characterizing} function $B$ is defined as follows. For any input $x\in\mathcal{X}$, the vicinity $B(x)$ is the local domain around (often centered at) $x$, and $B(x)\subset\mathcal{X}$. A vicinity $B(x)$ is often defined to be some $L^p$ norm of $x$ (where $p = 0, 1, 2, \infty$)~\cite{kurakin2016adversarial}, or domain-specific label-preserving transformations (e.g., tilting and zoom in/out)~\cite{athalye2018synthesizing,bhattacharya2019survey}. Specifically, a vicinity is characterised by a distance function $d$, and a predefined threshold $\epsilon$ as follows: $B(x_1) = \set{x_2\in\mathcal{X}  \mid  d(x_1,x_2) < \epsilon}$. Common notations of distances include
\begin{equation}\label{eq:d}
\begin{aligned}
    d(x_1, x_2) &= \norm{x_1 - x_2}_p, ~~~\mbox{(Additive in } L^p \mbox{ norm), or}\\
    d(x_1, x_2) &= 
    \begin{cases}
    \abs{\epsilon'}, &\mbox{if}~~f_\text{transform} (x_1, \epsilon') = x_2 ,\\
    \epsilon + 1, &\mbox{otherwise}
    \end{cases}  
\end{aligned}
\end{equation}
where the transform function mapping from $\mathcal{X}$ to $\mathcal{X}$ can be understood as a specific tranformation (\emph{e.g.}, whether an image is rotated or horizontally shifted) and its parameters ($\epsilon'$, \emph{e.g.}, the \textit{degree} of rotation).

Lastly, $\kappa$ is a constant threshold within the range of $[0, 1]$. When $\kappa = 0$ in \cref{eq:robustness}, it is commonly known as deterministic robustness~\cite{madry2017towards,pang2022robustness,li2023sok}. Otherwise, it is commonly known as probabilistic robustness~\cite{zhang2023proa,li2023sok}. We remark that it has been observed that completely eliminating adversarial examples (i.e., by having $\kappa = 0$) is often too stringent to be practical, compared to the alternative of keeping the possibility of undesirable events from occurring low (i.e., by having $\kappa$ slightly above zero)~\cite{zhang2023proa}.

\subsection{Robust Model Training}

To create robust models, various methods have been proposed, which have been reviewed in recent studies~\cite{silva2020opportunities,li2023sok}. State-of-the-art training methods can be broadly categorized into adversarial training~\cite{ganin2016domain}, certified training~\cite{singh2019abstract}, and a number of other approaches.

\paragraph{Adversarial Training}

Adversarial training is a widely-used and effective method for improving a model's empirical robustness. While there are many variants of the method, the most noticeable method involves solving the following optimization problem to achieve this goal.
\begin{equation}
\label{eq:adv_train}
    \min_h ~\operatorname{E}_{x\sim\mathcal{D}} \left[ \max_{t\in B(x)}\ \ell\big( h, t, G_t \big)\right]
\end{equation}
where $\ell$ is a suitably-chosen loss function (e.g., the 0-1, cross-entropy, or squared loss). The idea is to approximate the worst loss that can be induced by a perturbation for each training sample during training and optimize the parameter of model $h$ to improve the estimated worst-case robustness (in addition to standard accuracy).

A critical part of adversarial training is to search for adversarial inputs within the vicinity of the training samples. Goodfellow et al. introduce a fast gradient sign method (FGSM) to generate adversarial inputs~\cite{goodfellow2014explaining}. Adversarial training with FGSM significantly improves a model's robustness against adversarial samples generated through FGSM. Various other adversarial attacking methods are adopted for adversarial training as well. Among them, Projected Gradient Descent (PGD~\cite{madry2017towards}) based adversarial training is shown to be the most effective, in various domains, including image classification and reinforcement learning. In the context of large-scale image classification tasks, an ensemble adversarial training method further improves robustness through utilizing adversarial examples generated from multiple pre-trained models~\cite{kurakin2016adversarial}.

Despite the advancements made in adversarial training over the years, improving model robustness remains an open problem. This is partly due to the challenge posed by the trade-off between standard accuracy and robustness~\cite{tsipras2018robustness}. To this end, TRADES is proposed to balance this trade-off with a regularization term based on Kullback-Leibler (KL) divergence between the model's output on clean inputs and adversarial inputs~\cite{zhang2019theoretically}. This approach has achieved state-of-the-art performance on several benchmark datasets, including CIFAR-10. Nevertheless, a 15\%  accuracy drop is still observed.

More importantly, a significant limitation of adversarial training is that it does not certify a model's robustness against adversarial attacks~\cite{balunovic2020adversarial}. This lack of certification implies that the robustness of a model cannot be guaranteed, particularly as new and sophisticated adversarial attacking methods are being developed~\cite{athalye2018obfuscated}. For instance, it has been shown that a model trained through adversarial training remains vulnerable to new threats such as adaptive adversarial attacks~\cite{liu2019adaptiveface,tramer2020adaptive}. 
This limitation highlights the need for techniques that can provide certified robustness, i.e., a guarantee that the model is robust no matter what adversarial attacks are conducted.

\paragraph{Certified Training}

Certified training aims to train models that are certified to be robust~\cite{vaishnavi2022accelerating}. The idea is to soundly approximate the effect of any adversarial attack method and optimize the parameters of model $h$ so that the effect of any adversarial attack method is kept within a certain bound such that deterministic robustness is guaranteed~\cite{li2023sok}. The optimization problem is defined as follows. 
\begin{equation}
\label{eq:cert_train}
    \min_h ~\operatorname{E}_{x\sim\mathcal{D}} \left[ \sup_{t\in B(x),~c\neq G_t}\Big(\ell( h, t, G_t) - \ell( h, t, c \Big)\right]
\end{equation}
To soundly approximate the effect of any adversarial attack method, existing certified training methods use neural network verification techniques to soundly approximate the worst loss that can be induced by any perturbation within the vicinity of each training sample. If the label remains the same in the presence of such worst loss, the model is certified to be robust with respect to the sample. Note that after years of development, many neural network verification techniques have been proposed, e.g., \cite{zhang2018efficient,singh2019abstract,balunovic2020adversarial}.

Certified training methods however suffer from multiple shortcomings. First, they are computationally expensive. Although there has been a lot of development in neural network verification techniques, it is perhaps fair to say that such methods are still limited to relatively small neural networks. Given that certified training requires verifying the neural network robustness against each and every training sample, certified training is limited to small neural networks as of now.
Second, existing certified training methods often result in a significant drop in the model's clean accuracy, i.e., accuracy on clean, non-adversarial inputs~\cite{cohen2019certified, raghunathan2018certified}. The best clean accuracy achieved by certified training is typically 70\% of that from adversarial training on the CIFAR-10 dataset~\cite{tsipras2018robustness,shi2021fast}. Such dramatic accuracy drop makes their application in real-world systems rare as of now. Lastly, existing certified training methods usually only work for robustness defined based on the $L^p$ norms or in rare cases, simple transformation such as image rotation~\cite{cohen2019certified}.

\begin{figure}[t]
    \centering
    \includegraphics[width=0.9\linewidth]{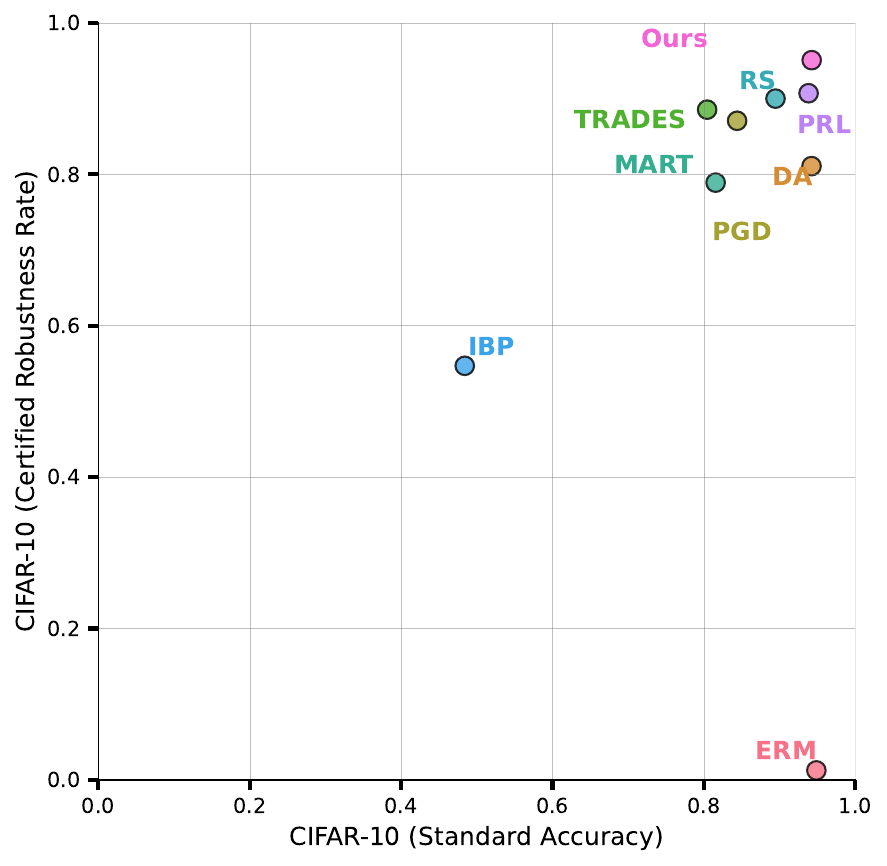}
    \caption{Comparing various methods using standard accuracy and certified robustness rates on CIFAR-10. Each circle represents one recently proposed method.}
    \label{fig:scatter}
\end{figure}

\subsection{Problem Definition}
In this work, we aim to develop a method that achieves certified probabilistic robustness whilst maintaining high accuracy. Unlike deterministic robustness which puts strict requirements on how a model should behave, probabilistic robustness relaxes the requirements by allowing a small of number of exceptions within the vicinity of a sample to have different labels, which makes it much more achievable in practice. Furthermore, certified probabilistic robustness provides theoretical guarantees for the model performance when faced with adversarial inputs, which could be useful for system-level decision making. In practice, it is often sufficient to keep the probability of undesirable events from occurring low.   

However, achieving high levels of certified probabilistic robustness and accuracy simultaneously model is challenging. The illustration in \cref{fig:scatter} highlights the delicate balance between maximizing clean accuracy and certified robustness that are achieved by state-of-the-art approaches (refer to Section~\ref{sec:evaluation} for details of these approaches). This work aims to address this challenge and provide a solution that meets the following criteria:

\begin{itemize}
\item Accuracy on clean data, i.e., the standard accuracy of the model evaluated using a test set that is disjoint from the training set;
\item Robustness under different attack scenarios the accuracy of the model concerning adversarial samples generated using state-of-the-art adversarial attacking methods such as Auto Attack~\cite{croce2020reliable};
\item Probabilistic certified robustness, i.e., the probability that samples in the testing set distribution that is certified to be robust;
\item Computational efficiency during both training and inference, to scale up to larger neural models and handle a large number of input cases efficiently;
\item Compatibility with existing architectures and frameworks making it easy to integrate and extend to suit specific use cases and applications.
\end{itemize}

\section{Our Method}

\begin{figure}[t]
  \centering
  \begin{subfigure}[b]{0.48\linewidth}
    \includegraphics[width=\linewidth]{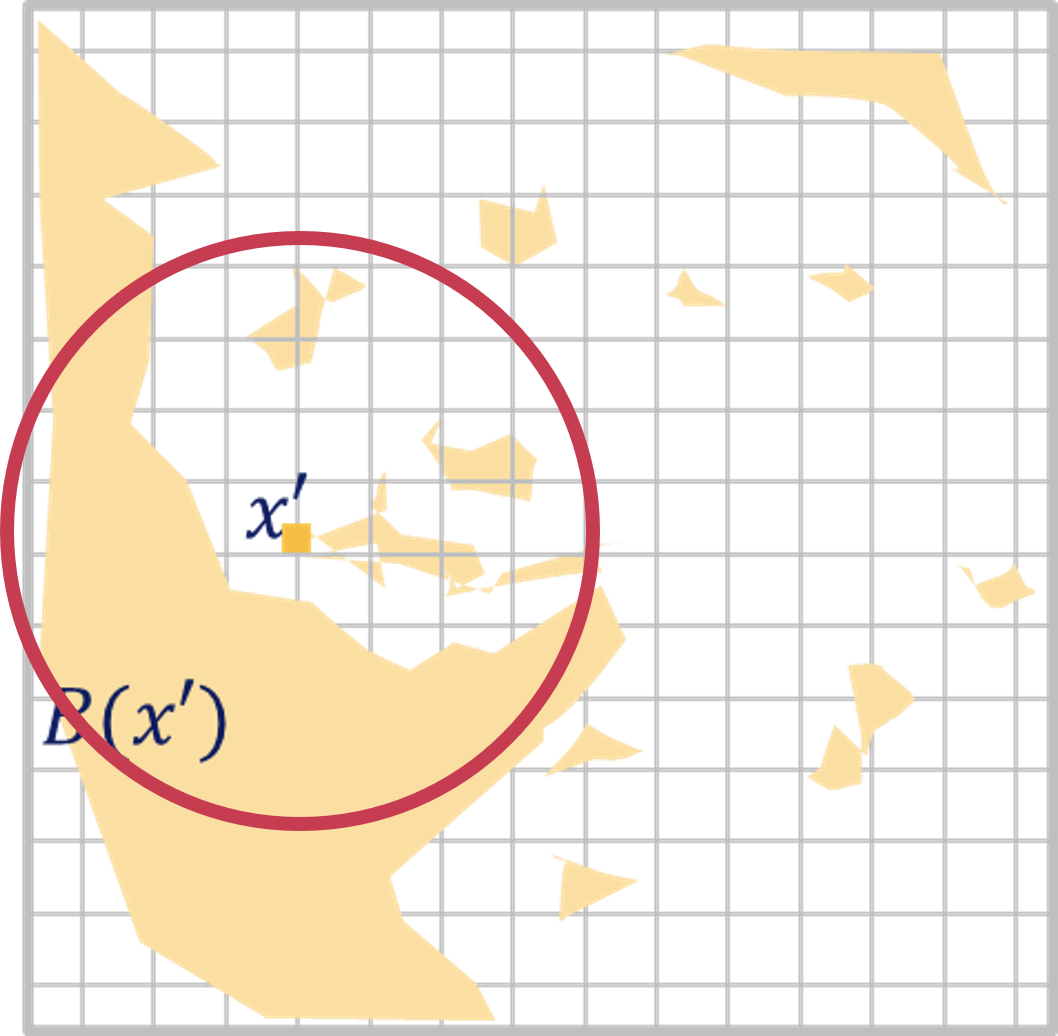}
    \caption{}
  \end{subfigure}
  \begin{subfigure}[b]{0.48\linewidth}
    \includegraphics[width=\linewidth]{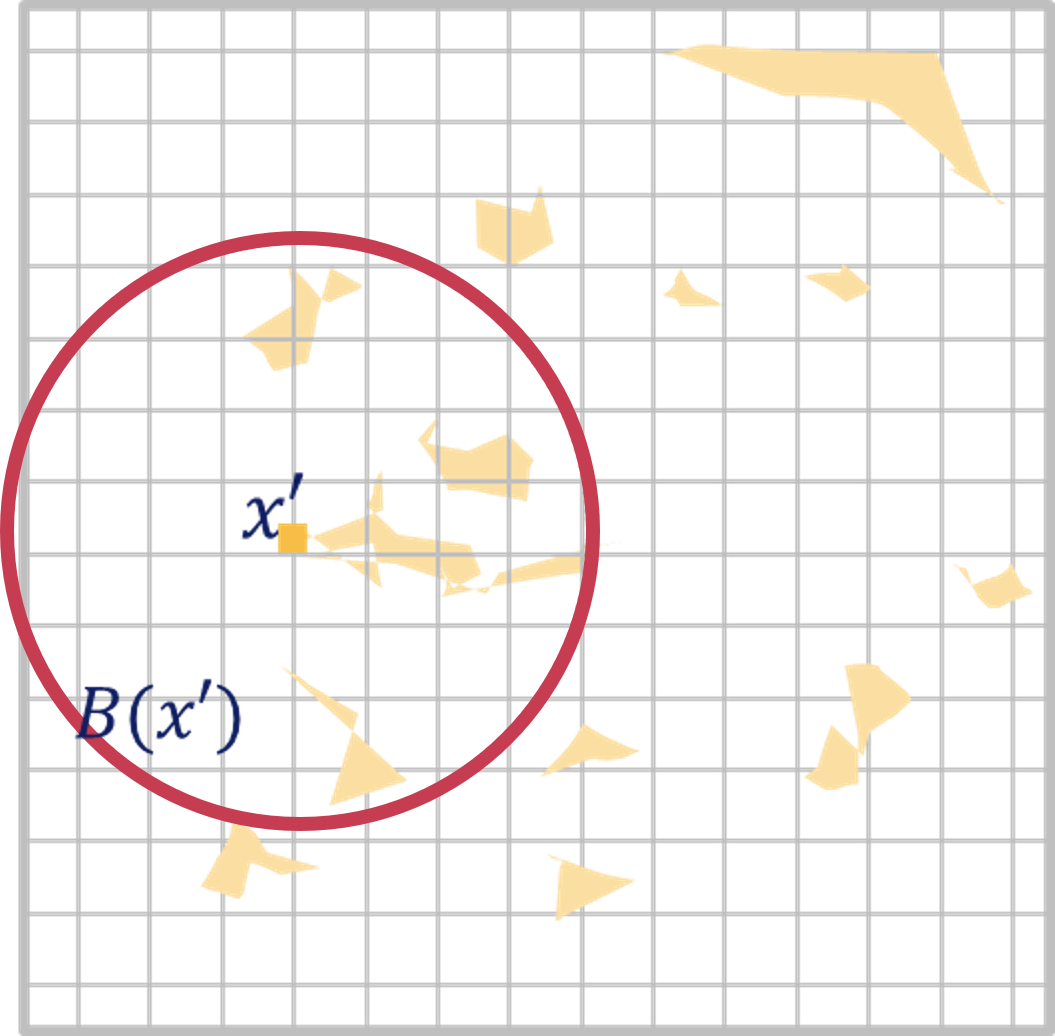}
    \caption{}
  \end{subfigure}
  \caption{Intuition of training: how can we achieve $P(t~\text{is correct} ~\mid t \in B(x')) \ge 1/2$? In a vicinity of any sample $x'$, (a) we may occasionally fail to achieve $P(t~\text{is correct} ~\mid t \in B(x')) \ge 1/2$; (b) $P(t~\text{is correct} ~\mid t \in B(x')) \ge 1/2$ always holds.}
  \label{fig:perturbation}
\end{figure}

Our method has two parts. The first is a training method which aims to improve probabilistic robustness (illustrated in \cref{fig:perturbation}) by minimizing the variance across the perturbation space for each sample in the training set. The second is an inference method which aims to establish certified robust prediction for a given sample. 

\subsection{Variance-Minimizing Training}
\label{sec:train}

To obtain a model that is both accurate and robust, we minimize the variance among model outputs for inputs within the same vicinity, alongside implementing empirical risk minimization. This training can be formulated as a Pareto optimization problem whose objective is as follows
\begin{equation}
\label[obj]{objective}
\begin{aligned}
    \min_h~~&\operatorname{E}_{x \sim \mathcal{D}}\Big[\operatorname{E}_{t\sim\mathcal{U}(B(x))}[\ell(h, t, G_t)] \Big]\\
    &\operatorname{E}_{x \sim \mathcal{D}} \Big[\operatorname{Var}_{t\sim\mathcal{U}(B(x)) }[\ell(h, t, G_t)]\Big]
\end{aligned}
\end{equation}
where the first term is essentially the objective of empirical risk minimization (ERM)~\cite{vapnik1999nature}, and the second term, variance of individual losses, is our novel objective.

Our goal is to minimize \cref{objective} by optimizing the weight parameters of neural network model $h$. The training algorithm to achieve this is presented in \cref{alg:train}, which outlines the specific operations involved. At each training step, we first sample a minibatch from the training data and for each (nominal) input in the minibatch, we sample a fixed number of (perturbed) inputs in the vicinity of the nominal input. Then, we use the neural network to make a prediction on each sample. Next, we compute the individual loss for each sample against the label of the given input independently. We then calculate the mean and standard deviation of these individual loss terms. Finally, we use a weighted sum of the mean and standard deviation as the effective loss function to back-propagate gradients and update the parameters of the neural network with the provided learning rate. This completes the flow of optimization. The presented algorithm is illustrated with stochastic gradient descent (SGD) optimizer but can be applied with other optimizers such as SGD with adaptive delta (Adadelta)~\cite{zeiler2012adadelta}.

In \cref{alg:train}, the loss function combines mean minimization and variance minimization, with a weighting factor $\lambda$ determining the importance of each component. Note that we use the square root of the variance term, allowing a linear combination of mean and standard deviation (SD) for the loss back-propagation.

\begin{algorithm}[tb]
\caption{Variance-Minimizing Training}
\label{alg:train}
\begin{algorithmic}[1]
    \STATE {\bfseries Input:} Training data $\set{(x_i, G_{x_i}) | i = 1,2, \ldots, k}\subset\mathcal{X}$, $L^\infty$ bound $\epsilon$, network architecture $h_\theta$ parametrized by $\theta$, step size $\eta$, sample size $n$, batch size $m$, and $\lambda$.
    \STATE {\bfseries Initialization:} Standard random initialization of $h_\theta$
    \STATE {\bfseries Output:} Robust network $h_\theta$
    
    \REPEAT
    \STATE Uniformly sample $\set{(t_i, G_{t_i}) | i = 1,2, \ldots, m}$, a minibatch of training data where $m<k$
    \FOR{$i = 1,2, \ldots, m$}
        \STATE Draw $\set{\tau_j| j = 1, 2, \ldots, n} \sim \mathcal{U}(t_i - \epsilon, t_i + \epsilon)$ where $\mathcal{U}$ is a uniform distribution
        \FORALL{$j=1, 2, \ldots, n$}
            \STATE $u_j \gets l_\text{Cross-entropy}(h_\theta(\tau_j), G_{t_i})$
        \ENDFOR
        \STATE $\mu_i \gets \sum_{j=1}^n u_j / n$
        \STATE $\sigma_i \gets(\sum_{a=1}^n\sum_{b=1}^n(u_a - u_b)^2 / n)^{1/2}$
    \ENDFOR
    \STATE{$\theta\gets \theta-\eta \sum_{i=1}^m \nabla_\theta[\mu_i + \textcolor{catalina_blue}{\lambda}~\sigma_i]/m$}

    \UNTIL{convergence}
\end{algorithmic}
\end{algorithm}

Intuitively, our \cref{objective} allows us to improve the model's robustness without depending on any specific adversarial attacking methods. Instead, we improve model robustness by minimizing the spread (standard deviation) of model prediction alongside the traditional ERM method. Random sampling is adopted and the adversarial attack in each training step is avoided.

In the ideal case, if, for a given $x\in\mathcal{X}$ and model $h$, a variable in the vicinity around $x$ is correctly predicted by $h$ and prediction of \emph{any pair} of samples in this vicinity is the same, then $h$ achieves deterministic robustness in that vicinity. In the more likely case, the variance of the loss within the vicinity of each training sample is minimized through training and as a result, many of the samples within the vicinity may have the same (correct) prediction. Note that, unlike existing adversarial training methods which either rely on pre-computed adversarial examples~\cite{miyato2018virtual} or adversarial examples generated during training~\cite{zhang2019theoretically} (often paying a high training cost), our training method is independent of specific attacking methods.

Both terms in \cref{objective} are crucial to improving the robustness of the model. Variance in the data represents the difference between individual observations. High variance means that the observations are scattered, while low variance means they are tightly clustered around the mean. Therefore, we believe that reducing the variance of the predictions can make the model more robust. Conversely, minimizing the mean alone (such as data augmentation~\cite{wen2020time}) leaves the outliers of the distribution to be unpredictable, which can lead to the existence of adversarial examples.

Formally, we want to show why models with lower variance among nearby predictions are more robust.
\begin{proposition}\label{prop:quantile}
If two distributions with the same mean have different variances, where the variance of one is less than the other, then for any quantile level $q$ in the range $0 < q < 1$, the upper bound of the $q$-th quantile of the distribution with the lower variance is less than the upper bound of the $q$-th quantile of the distribution with the higher variance.

\end{proposition}
\begin{proof}
We start with Chebyshev's inequality. Chebyshev’s inequality provides an upper bound on the tail probabilities of a random
variable based on its variance. Let $Z$ (integrable) be a random variable with finite expected value $\mu$ and finite non-zero variance $\sigma^2$. Then for any real number $\lambda > 0$,
\begin{equation}
    P\big(\abs{Z-\mu} \geq \lambda\sigma \big)\leq \frac {1}{\lambda^{2}}
\end{equation}
which states that for any probability distribution, the proportion of data within $\lambda$ standard deviations of the mean is at least $1-1/\lambda^2$, and we can further derive:
\begin{equation}
\begin{aligned}
    P\big(\abs{Z-\mu} \geq \lambda\sigma \big) = P\big(Z-\mu > \lambda\sigma \big)&\leq \frac {1}{\lambda^{2}}\\
    P\big(Z \leq \lambda\sigma +\mu\big)&\geq 1 - \frac {1}{\lambda^{2}}    
\end{aligned}
\end{equation}
Let $\lambda\sigma =  \mu - z$, we can have the inequality flipped like:
\begin{equation}
    P\big(Z \leq z\big)\geq 1 - \frac {\sigma^2}{(z - \mu)^2}
\end{equation}
For any given $z$, When the variance $\sigma^2$ decreases, the lower bound for $P\big(Z \leq z\big)$ increases. Hence, minimizing the variance is essentially reducing the probability of examples far away from the mean.
\end{proof}

Although a higher $\lambda$ may appear more desirable, as it covers a larger vicinity, the model may be tuned to prioritize reducing variance over the mean, as our experiments show. If we omit the spread term in the loss function, the model minimizes expectation, similar to training with augmented data~\cite{wen2020time}. Omitting the expectation term is not recommended as it can lead to a model which makes poor predictions for all samples.

\subsection{Inference and Certifying}
\label{sec:certify}

The second part of our approach is an inference algorithm which aims to provide certified probabilistic robustness when possible. 
According to \cref{eq:robustness}, to establish certified probabilistic robustness, we must show that there is a guaranteed upper bound on the probability of adversarial examples, i.e., some threshold $\kappa$. Intuitively, we would like to know for sure that among all the samples within the vicinity around an input, at least $1-\kappa$ of them are not adversarial examples.

Our inference method aims to certify the robustness, i.e., while providing prediction to an input variable, our inference method also offers certified probabilistic robustness as described above. To present this inference method, we first demonstrate our algorithm and then explain how it provides certified robustness and illustrate the difference between inference with certified robustness and vanilla inference.

\paragraph{Algorithm}

The general idea of our inference method is captured below. For any $x\in\mathcal{X}$ and model $h$,
\begin{equation}
    h^*(x) \coloneqq (h*B)(x) \coloneqq \int _\mathcal{X}h(\tau )\operatorname{\mathbb{I}}(x-\tau\in B(x))\,d\tau
\end{equation}
where symbol $*$ denotes convolution, the mathematical operation on two functions. A superscript on $h$ indicates that the model is based on the proposed inference instead of the ordinary inference. $\operatorname{\mathbb{I}}(\phi)$ a function that returns 1 if $\phi$ is satisfied and 0 otherwise. A more feasible step-by-step implementation of this idea is presented in \cref{alg:inference}

\begin{algorithm}[tb]
\caption{Inference With Certified Robustness}
\label{alg:inference}
\begin{algorithmic}[1]
    \STATE {\bfseries Input:} Test data $\set{(x_i, G_{x_i}) | i = 1,2, \ldots, k}\subset\mathcal{S}$, $L^p$ bound $\epsilon$, model (network) $h$, threshold $\kappa$, (statistical)  significance level $\alpha$
    \STATE {\bfseries Initialization:} Certified Robustness \textit{rate} $c\gets 0$
    \STATE {\bfseries Output:} Model prediction $\set{(x_i, \hat{y}_{x_i}) | i = 1,2, \ldots, k}$ for each case in test data; certified robustness for model $h$ on each case in test data against $L^p$-adversary at bound $\epsilon$; certified robustness \textit{rate} for model $h$.
    
    \FOR{$i = 1,2, \ldots, k$}
        \STATE Predictions $S \gets$ empty dictionary
        \REPEAT
            \STATE Sample $\tau \sim \mathcal{U}(\set{t\in\mathcal{X}|\norm{t - x_i}_p \leq \epsilon})$ where $\mathcal{U}$ is a uniform distribution
            \STATE Prediction $s\gets\argmin \ell(h, \tau, c)$
            \IF{$s$ in $S$}
                \STATE $S[s] \gets S[s] + 1$
            \ELSE
                \STATE $S[s] \gets 0$
            \ENDIF
            \STATE Most likely prediction $u = \argmax S$
            \STATE Highest count $v = S[u]$
            \STATE Total count $w = \sum_j S[j]$
            \STATE $p_\text{left} \gets \operatorname{Left-binomial-test}(v, w, p_0 = 1 - \kappa)$
            \STATE $p_\text{right} \gets \operatorname{Right-binomial-test}(v, w, p_0 = 1 - \kappa)$
        \UNTIL{$p_\text{left}<\alpha$ or $p_\text{right}<\alpha$}
        \IF{$p_\text{left}<\alpha$}
            \STATE Inference on $x_i$ is without certified robustness
        \ELSE
            \STATE Inference on $x_i$ has certified robustness
            \STATE $c\gets c + 1 / k$
        \ENDIF
        \STATE Prediction $\hat{y}_{x_i} \gets u$ for $x_i$
    \ENDFOR
\end{algorithmic}
\end{algorithm}

The intuition of \cref{alg:inference} is that to make a prediction for any input variable $x$, we always sample sufficiently many samples in the vicinity of $x$ and make the prediction based on the majority of the predictions. The idea is that in this way, a model would only make mistakes when more than half of the sampled samples are predicted wrongly. It is important to note that this inference method alone, can only moderately improve the standard accuracy or adversarial accuracy (refer to \cref{subsec:rq} for our ablation study results). This is because the samples within the vicinity of an adversarial example are often likely to be predicted wrongly as well. The above-proposed inference method works effectively with our training method as the model is trained to make similar predictions within a vicinity.

We reduce the task of verifying probabilistic robustness to determining the probability of correct predictions using ordinary inference. Next, to determine the probability of correct predictions using the ordinary inference, we adopt an established method known as the exact binomial test~\cite{blitzstein2019introduction}.

\paragraph{Exact Binomial Test}

The Binomial test is a statistical procedure used to test a hypothesis about the population proportion of a binary variable based on a sample of observations. It can be used to determine whether the proportion of one level in a binary variable is less than, greater than, or not equal to a specific claimed value. To evaluate the hypothesis that the proportion of a certain class of prediction around an input is higher than $\kappa$, e.g., 10\%, we conduct a binomial test using sample data. We use the following formula to calculate the probability of obtaining the observed occurrence of this class, or more extremely if the true proportion is less than 10\%.
\begin{equation*}
P(Z \geq z \mid p = \kappa) = 1 - \sum_{i=0}^{z-1} \binom{n}{i} (\kappa)^i (1-\kappa)^{n-i}
\end{equation*}
where $Z$ is the number of observed occurrences of this class in the sample; $z$ is the observed number of occurrences of this class; $n$ is the sample size; $p$ is the claimed population proportion (in this case, 0.1); $\binom{n}{i}$ is the binomial coefficient, which calculates the number of ways to choose $i$ items from a set of $n$ items.

If the resulting probability is less than a pre-determined significance level (e.g. $\alpha=0.05$), we reject the null hypothesis that the proportion of occurrences is greater than or equal to 10\% and conclude that it is lower. Otherwise, we fail to reject the null hypothesis and conclude that there is not enough evidence to suggest that the proportion is lower than 10\%.

In \cref{alg:inference}, we perform both left-tail (i.e., $P(Z \le z)$) and right-tail binomial tests to ensure that we can either accept that the probability is greater than $1-\kappa$ or less than $1-\kappa$. This provides certainty as to whether the prediction on the test case is certified as robust or not. The level of statistical significance is determined by $\alpha$. As $\alpha$ decreases, the statistical significance increases, which means that the certification is less likely to result in a false positive. Additionally, those cases that are not certified as robust have a lower likelihood of being false negatives. Although $\kappa$ and $\alpha$ are typically selected within the range of $10^{-1}$ to $10^{-4}$, decreasing both values, i.e., $\kappa\to 0$ and $\alpha\to 0$, can make the certification more reliable.

We use sequential sampling to obtain the required sample size at runtime, which has proven to be optimal~\cite{wald1992sequential}. We stop collecting data once the probability of either the right or left tail crosses a predefined false positive rate. We make a decision based on which tail has crossed the threshold and certify the prediction as either robust or non-robust accordingly. The binomial test is described in detail in lines 7-21 of \cref{alg:inference}.

\begin{theorem}
\label{thm:sample}
Let $x$ be a sample. If Algorithm 2 returns that $x$ has certified robustness, i.e., $p_\text{right} < \alpha$, then the probabilistic robustness of $x$ is greater than $1-\kappa$ is satisfied. 
\end{theorem}

\section{Experiment} \label{sec:evaluation}

In the following, we systematically evaluate our method by answering multiple research questions.

\subsection{Experimental Setting}
\paragraph{Datasets}

Experiments are run on widely-used classification datasets: MNIST~\cite{lecunmnist}, SVHN~\cite{netzer2011reading}, CIFAR-10~\cite{krizhevsky2009learning}, and CIFAR-100~\cite{krizhevsky2009learning}. The details of these datasets for our study on  robustness of classification models~\cite{madry2017towards,tramer2017ensemble,kurakin2018adversarial} are present in \cref{tab:datasets}. In brief, the SVHN, CIFAR-10, and CIFAR-100 datasets consist of 32$\times$32 color images, while the MNIST dataset comprises 28$\times$28 grayscale images. The original training set of each dataset comprises a minimum of 50,000 samples, which are partitioned into training and validation sets following a ratio of 8:2.

\begin{table}[t]
\centering
\caption{Details on image classification datasets, and perturbation bounds for each task}
\label{tab:datasets}
\footnotesize
\begin{tabular}{@{}l|cccc@{}}
\toprule
Task & MNIST & SVHN & CIFAR-10 & CIFAR-100 \\
\midrule
Training Images & 48,000 & 58,606 & 40,000 & 40,000 \\
Validation Images & 12,000 & 14,651 & 10,000 & 10,000 \\
Testing Images & 10,000 & 26,032 & 10,000 & 10,000 \\
Image size & \cellcolor{lightgray!25}$28\times28$ & \multicolumn{3}{c}{\cellcolor{lightgray!10}$32\times32$}\\
Color Channels & \cellcolor{lightgray!25}1 & \multicolumn{3}{c}{\cellcolor{lightgray!10}3}\\
Classes & \multicolumn{3}{c}{\cellcolor{catalina_blue!10}10} &  \multicolumn{1}{c}{\cellcolor{lightgray!10}100} \\\midrule
$L^\infty$ bound & \cellcolor{lightgray!25} 0.1 or 0.3 & \multicolumn{3}{c}{\cellcolor{lightgray!10}2/255 or 8/255}  \\
Translation & \multicolumn{4}{c}{\cellcolor{lightgray!10}$\pm 0.3$} \\
Rotation & \multicolumn{4}{c}{\cellcolor{lightgray!10}$\pm35\degree$} \\
Scaling Factor & \multicolumn{4}{c}{\cellcolor{lightgray!10}$\pm 0.3$} \\
\bottomrule
\end{tabular}
\end{table}

\paragraph{Model Architectures}
We adopt multiple model architectures to train the classifiers on the above-mentioned datasets. The details of these architectures are summarized in \cref{tab:architectures}. These architectures all have been studied by exisiting robustness improving methods, as shown in the \emph{Works} column. In short, the model size ranges from 378,562 parameters for the small CNN7 model, to 11,689,512 parameters for the more complex ResNet-18 model. 

\begin{table}[t]
\centering
\caption{Details of model architectures}
\footnotesize
\begin{tabular}{@{}l|c|c@{}}
\toprule
Model & \# Parameters & Works \\ 
\midrule
        ResNet-18~\cite{he2016deep,zhang2019theoretically} & 11,689,512 & \cite{zhang2019theoretically,wang2020improving,robey2022probabilistically} \\ 
        Wide-ResNet-8~\cite{zagoruyko2016wide} & 3,000,074 & \cite{shi2021fast} \\ 
        CifarResNet-110~\cite{he2016deep} & 1,730,474 & \cite{cohen2019certified} \\ 
        CNN7 & 378,562 & \cite{shi2021fast} \\ 
        Basic ConvNet & 1,663,370 & \cite{zhang2019theoretically,wang2020improving,robey2022probabilistically} \\ 
\bottomrule
\end{tabular}
\label{tab:architectures}
\end{table}

\begin{table}[t]
\centering
\caption{Compatibility between different methods and model architectures}
\footnotesize
\begin{tabular}{l| ccccc}
\toprule
Approach & \begin{tabular}[c]{@{}c@{}}ResNet \\-18\end{tabular} & \begin{tabular}[c]{@{}c@{}}Wide- \\ResNet\\-8\end{tabular} & \begin{tabular}[c]{@{}c@{}}Cifar\\ResNet \\-110\end{tabular} & CNN7 & \begin{tabular}[c]{@{}c@{}}Basic \\ConvNet\end{tabular}  \\
\midrule
        ERM & \checkmark & \checkmark& \checkmark& \checkmark& \checkmark \\ 
        DA & \checkmark& \checkmark& \checkmark& \checkmark& \checkmark \\ 
        PGDT & \checkmark& \checkmark& $\times$& $\times$& \checkmark \\ 
        TRADES & \checkmark& \checkmark& $\times$& $\times$& \checkmark \\ 
        MART & \checkmark& \checkmark& $\times$& $\times$& \checkmark \\ 
        RS & $\times$ & $\times$& \checkmark& $\times$& \checkmark \\ 
        IBP & $\times$& \checkmark& $\times$& \checkmark& \checkmark \\ 
        PRL & \checkmark& \checkmark& \checkmark& \checkmark& \checkmark \\ 
        Ours & \checkmark& \checkmark& \checkmark& \checkmark& \checkmark \\ 
\midrule
\rowcolor{mygray}\multicolumn{6}{l}{\begin{tabular}[c]{@{}l@{}}The Basic ConvNet architecture is suitable exclusively for the MNIST\\ dataset, while tasks involving SVHN and CIFAR datasets require\\ the utilization of residual networks with more parameters. \end{tabular}}\\
\bottomrule
\end{tabular}
\label{tab:compatibility}
\end{table}

\begin{table}[t]
\centering
\caption{Effectiveness metrics. To evaluate a model $h$ on any input data from some distribution $\mathcal{D}$, we assume that a test set $\mathcal{S}$ generalises the distribution, and $|\mathcal{S}|$ is the number of testing samples.
}
\label{tab:metrics}
\footnotesize
\begin{tabularx}{\linewidth}{p{4em}|X|X}
\toprule
Metric & Formula & Meaning \\ \midrule
Standard Accuracy & $\frac{1}{|\mathcal{S}|}\sum_{(x, G_x)\in \mathcal{S}}$   $\quad\Big(\operatorname{\mathbb{I}}\big(\argmax h(x) $  $= G_x \big)\Big) $ & The probability that the model's prediction is correct for an input from the data distribution $\mathcal{D}$. \\ \midrule
Certified Robustness Rate & $\frac{1}{|\mathcal{S}|}\sum_{(x, ~\textcolor{mygray}{\_})\in \mathcal{S}} $$\Big(\operatorname{\mathbb{I}}\big(h(x) $$~\text{\small is with}$  $ \text{\small certified robustness}\big)\Big)$ & The probability that the model's prediction has certified robustness for an input from the data distribution $\mathcal{D}$. \\ \midrule
Certified Robust Accuracy & $\frac{1}{|\mathcal{S}|}\sum_{(x, G_x)\in \mathcal{S}} \Big(\operatorname{\mathbb{I}}\big(h(x) $$~\text{\small is with}$  $ \text{\small certified robustness}\big) \times \operatorname{\mathbb{I}}\big(\argmax h(x) = G_x\big)\Big)$ & The probability that the model's prediction has certified robustness and this prediction is correct, for an input from the data distribution $\mathcal{D}$. \\ \midrule
Defence Success Rate & $\frac{1}{|\mathcal{S}|}\sum_{(x, G_x)\in \mathcal{S}} $   $\quad\Big(\operatorname{\mathbb{I}}\big(\argmax h(A(h, x, G_x)) = G_x\big)\Big)$ & The probability that the model's prediction is correct when the input has been perturbed by adversarial attack $A$, for an input from the data distribution $\mathcal{D}$. \\ \midrule
\rowcolor{mygray}\multicolumn{3}{l}{$\operatorname{\mathbb{I}}(\phi)$ a function that returns 1 if $\phi$ is satisfied and 0 otherwise}\\\bottomrule
\end{tabularx}
\end{table}

\paragraph{Baselines}

In the evaluation, we compare our method with eight baselines: 1) Empirical Risk Minimization (ERM)~\cite{vapnik1999nature} is the standard training approach without any additional modifications; 2) Data Augmented training (DA)~\cite{shorten2019survey} trains the model with the samples augmented by applying various transformations to improve the generalization and robustness; 3) PGD-Training (PGDT)~\cite{madry2017towards} aims to optimize the model’s prediction error for the training samples and their surrounding neighborhoods, which are sampled by Projected Gradient Descent (PGD); 4) TRADES~\cite{zhang2019theoretically} attempts to minimize both the prediction error of the original samples and the prediction inconsistency between them and their neighborhoods; 5) MART~\cite{wang2020improving} improves TRADES by a specific focus on misclassified samples; 6) Randomized Smoothing (RS)~\cite{cohen2019certified} provides certified robust accuracy by adding noise to inputs during training; 7) IBP~\cite{shi2021fast} acquires the tractable upper bound for the worst-case perturbation and then provides a deterministic certificate of robustness; 8) PRL~\cite{robey2022probabilistically} is a probabilistic training method that aims to reduce the proportion of adversarial examples. The implementations of all these baselines are obtained from their respective original repositories.

These baselines all aim at a robust and accurate model, although they originally pursue different metrics. PGDT~\cite{madry2017towards}, TRADES~\cite{zhang2019theoretically}, and MART~\cite{wang2020improving} are adversarial training that aims for empirical robustness. For certified training, e.g., IBP~\cite{shi2021fast}, the emphasis is placed on theoretical guarantees. Consequently, the primary objective is to optimize in a way such that no adversarial examples exist in the vicinity of each data point. This pursuit of adversarial robustness takes precedence over maintaining high accuracy levels, if necessary. PRL~\cite{robey2022probabilistically} aims to minimize the proportion of adversarial examples (based on training for probabilistic robustness). Their approach improves model robustness by maximizing the lower bound of the probability that the model's predictions are correct under a certain level of perturbation.

Note that not all methods can be applied all model architectures. \cref{tab:compatibility} summarizes the compatibility between the methods and model architectures. It should be noted that our method, along with ERM, DA, and PRL, applies to all architectures. We systematically evaluate each method for each model architecture to find the best-suited architecture for each method and dataset, e.g., TRADES~\cite{zhang2019theoretically} eventually finds that ResNet-18 is the best matching architecture for SVHN, and the basic ConvNet for MNIST, while being not compatible to CNN7 (refer to \cref{tab:compatibility}).  The most suitable architecture for each approach on different tasks is as follows: 1) For the MNIST dataset, all approaches except IBP can utilize the basic ConvNet architecture, while IBP adopts CNN7. 2) For the SVHN or CIFAR-10/100 datasets, all approaches except IBP or RS can use ResNet-18, while IBP utilizes Wide-ResNet-8 and RS adopts CifarResNet-110. In the following, we report the experimental results according to the most suited architecture.

\paragraph{Reproducibility}
We provide our code implementation, trained models, and supplementary materials on our repository at \url{https://github.com/soumission-anonyme}. In our training, we use different optimization strategies for different benchmarks to obtain the best performance. For example, we use Adadelta optimizer~\cite{zeiler2012adadelta} with a learning rate of 1.0 for 150 epochs to optimize Basic ConvNet on MNIST. For the other three tasks, we use the SGD optimizer with an initial learning rate of 0.01 and weight decay of 3.5e-3. The learning rate for SGD is reduced by a factor of 10 at epochs 55, 75, and 90. Our experiments are conducted on a server with an x86\_64 CPU featuring 8 cores running at 3.22GHz, 54.93GB of RAM, and an NVIDIA RTX 2080Ti GPU with 11.3 GB of memory.

\begin{table*}[t]
\centering
\caption{The certified robust accuracy, certified robustness rate, and standard accuracy of different approaches on various datasets. The table provides insight into the effectiveness of each approach in achieving high levels of certified robustness and accuracy.}
\footnotesize
\begin{tabular}{@{}l| cccc| cccc| cccc@{}}
\toprule
Approach & \multicolumn{4}{c}{Standard Accuracy} & \multicolumn{4}{c}{Certified Robustness Rate} & \multicolumn{4}{c}{Certified Robust Accuracy} \\
\cline{2-5} \cline{6-9} \cline{10-13}
~ & \scriptsize CIFAR-100 & \scriptsize CIFAR-10 & \scriptsize SVHN & \scriptsize MNIST & \scriptsize CIFAR-100 & \scriptsize CIFAR-10 & \scriptsize SVHN & \scriptsize MNIST & \scriptsize CIFAR-100 & \scriptsize CIFAR-10 & \scriptsize SVHN & \scriptsize MNIST \\
\midrule
        ERM~\cite{vapnik1999nature} & \textbf{81.03} & \textbf{94.85} & 94.44 & 99.37 & 9.28 & 1.25 & 52.72 & 26.01 & 4.52 & 1.25 & 51.04 & 24.96 \\ 
        DA~\cite{shorten2019survey} & 78.27 & 94.21 & 94.69 & \textbf{99.42} & 15.04 & 81.08 & 82.08 & 85.23 & 6.15 & 76.07 & 82.01 & 84.12 \\ 
        PGDT~\cite{madry2017towards} & 64.35 & 84.38 & 91.19 & 99.16 & 57.07 & 87.07 & 87.89 & 94.65 & 32.93 & 82.90 & 86.68 & 94.63 \\ 
        TRADES~\cite{zhang2019theoretically} & 62.55 & 80.42 & 86.16 & 99.10 & 59.27 & 88.54 & 87.89 & 94.76 & 38.85 & 78.80 & 84.76 & 94.61 \\ 
        MART~\cite{wang2020improving} & 63.68 & 81.54 & 90.20 & 98.94 & 58.79 & 78.90 & 85.23 & 94.13 & 49.37 & 72.21 & 78.82 & 94.09 \\ 
        RS~\cite{cohen2019certified} & 56.87 & 89.45 & 88.35 & 97.16 & 60.38 & 90.00 & 76.29 & 87.15 & 47.50 & 87.98 & 70.64 & 86.29 \\ 
        IBP~\cite{shi2021fast} & 39.45 & 48.40 & 73.09 & 97.78 & 49.34 & 54.70 & 61.94 & 89.18 & 29.20 & 40.00 & 57.26 & 88.51 \\ 
        PRL~\cite{robey2022probabilistically} & 64.89 & 93.82 & 92.00 & 99.32 & 56.71 & 90.71 & 93.11 & 96.03 & 50.77 & 90.63 & 91.07 & 95.01 \\ 
        Ours & 65.56 & 94.23 & \textbf{94.79} & 99.32 & \textbf{62.05} & \textbf{95.08} & \textbf{93.15} & \textbf{97.80} & \textbf{52.07} & \textbf{91.75} & \textbf{92.81} & \textbf{96.80} \\ 
\midrule
\rowcolor{mygray}\multicolumn{13}{l}{$\kappa = 10^{-2}, 1 - \alpha = 0.99$; $L^\infty$ bound at 0.3 for MNIST, and 8/255 for CIFAR-10, CIFAR-100, or SVHN. }\\
\bottomrule
\end{tabular}
\label{tab:certify}
\end{table*}

\subsection{Research Questions and Answers}
We seek to answer the following research questions through our experiments. 
\label{subsec:rq}

\paragraph{RQ1: Is our method effective in achieving robustness whilst maintaining accuracy?}

To answer this question, we evaluate the performance of our method and baseline methods using multiple metrics such as standard accuracy, certified robustness rate, and certified robust accuracy, as defined in \cref{tab:metrics}. 
Note that for the baseline methods that do not inherently report probabilistic certified robustness, we run the exact binomial test to verify their certified robustness rate. The results are shown in \cref{tab:certify}. 

In terms of certified robust accuracy, it can be observed that our method has the highest certified robust accuracy on all four datasets, with an average value of 83.36\%, which means that our method best strikes the balance between accuracy and robustness. In comparison, PRL is the best performing baseline method, with an average certified robust accuracy of 81.87\%. Furthermore, comparing the results on the different datasets, we observe that our method outperforms PRL more when the dataset is more complex. Additionally, 
the average certified robust accuracy of the best adversarial training method (i.e., PGDT) is 74.29\%, which is 11.42\% lower than ours. Randomized smoothing and IBP yield even lower results at 73.10\% and 53.74\% respectively, although they still outperform normal training (i.e., ERM) whose certified robust accuracy is only 20.44\%.

Referring to the provided definitions in \cref{tab:metrics}, it is evident that achieving a high certified robust accuracy necessitates both a high standard accuracy and a high certified robustness rate. In the following, we compare our method and baseline methods based on these two metrics separately. 

In terms of standard accuracy, our method exhibits a reasonably small sacrifice on standard accuracy while achieving robustness, compared to most of the existing methods. On the CIFAR-10, SVHN, and MNIST datasets, our method has a slight decreased accuracy in comparison to ERM, with a maximum reduction of less than 0.7\% and an average value of 0.1\% that closely approached DA. On CIFAR-100, although there is a noticeable decrease in accuracy compared to ERM, our method still ranked as the second-best training approach, only surpassed by DA. In addition, 
adversarial training results in a minimum 8.35\% drop in accuracy, certified training usually leads to over 40\% accuracy drop, and randomised smoothing results in a 10.31\% accuracy drop. These baselines sacrifice standard accuracy for their respective training objective. For example, randomised smoothing introduces Gaussian noise during the training process to improve the model's robustness to perturbations. However, it can inadvertently push some of the original samples farther away from their true labels, leading to a reduction in accuracy.

In terms of certified robustness rate, our method achieves the best performance on certified robustness rate, with an average value of 87.02\%, which is 3.42\% and 36.42\% higher than probabilistically robust learning and certified training, respectively. The baselines which has higher accuracy than ours, i.e., ERM and DA, are significantly lower certified robustness rate, with an average rate of 22.32\% and 65.86\%, respectively. \\

\begin{table}[t]
\centering
\caption{The certified robust accuracy of different approaches on CIFAR-10 with respect to different transformations.}
\footnotesize
\begin{tabular}{l| c|c|c|c}
\toprule
\cmidrule(lr){2-3}\cmidrule(lr){4-5}
Approach & \begin{tabular}[c]{@{}c@{}}Translation \\($\pm0.3$)\end{tabular} & \begin{tabular}[c]{@{}c@{}}Rotation \\($\pm35\degree$)\end{tabular}   & Affine & \begin{tabular}[c]{@{}c@{}}Scale \\($\pm0.3$)\end{tabular}  \\
\midrule
        ERM & 92.85 & 93.95 & 92.85 & 93.01 \\ 
        DA & 93.97 & 92.76 & 92.75 & 93.23 \\ 
        PGDT & 64.35 & 74.23 & 61.34 & 69.46 \\ 
        TRADES & 68.72 & 74.89 & 64.56 & 74.45 \\ 
        MART & 73.48 & 81.35 & 74.23 & 68.49 \\ 
        RS & 87.25 & 86.29 & 82.51 & 87.58 \\ 
        IBP & 46.52 & 46.94 & 44.41 & 47.80 \\ 
        PRL & 90.68 & 91.74 & 89.23 & 90.92 \\ 
        Ours & \textbf{93.28} & \textbf{94.15} & \textbf{93.28} & \textbf{93.25} \\ 
\midrule
\rowcolor{mygray}\multicolumn{5}{l}{$\kappa = 10^{-2}, 1 - \alpha = 0.99$ }\\
\bottomrule
\end{tabular}
\label{tab:transform}
\end{table}

\noindent \textbf{More than $L^p$ transformation.} While the existing robustness certification methods such as certified training and randomized smoothing primarily focus on $L^p$ transformations of images~\cite{shi2021fast}, as presented in \cref{tab:certify}, we are also interested in the certified robustness on other transformations, such as translation, rotation, affine, and scaling. In our experiments, we randomly perturb the input within the given range of each transformation and report the corresponding certified robust accuracy. Similar to the previous experiments, we apply the exact binomial test to verify the robustness of the model obtained by different training algorithms with respect to non-$L^p$ norm transformation. The hyper-parameters $\kappa$ and $\alpha$ are set as $10^{-2}$ and $0.01$, respectively. We present the results on CIFAR-10 in \cref{tab:transform} and similar results are obtained on other datasets. It can be observed that our method consistently achieves the highest certified robust accuracy across all transformations, surpassing the threshold of 93.49\%. Combined with the results shown in \cref{tab:certify}, it shows that our method is the most robust training algorithm against different kinds of perturbations, including both $L^p$ and non-$L^p$ transformations. The second highest is DA, with all results above 92.23\%. This is likely because, rotation, translation, and scaling are frequently used in data augmentation. Remarkably, ERM achieves the third-highest certified robust accuracy, which can be attributed to the inherent robustness of convolutional layers to these non-$L^p$ transformations. This robustness is due to their ability to capture and extract local patterns and spatial relationships in images through shared weights, local receptive fields, and spatial pooling operations~\cite{he2016deep}. 
In addition, PRL has slightly worse performance than our method, i.e., by 3.05\%.
\begin{center}
\fcolorbox{black}{white!20}{\parbox{0.97\linewidth}
    {
        \emph{\textbf{Answer to RQ1}}:
        The proposed approach achieves the highest certified robust accuracy (83.4\%), primarily due to its highest certified robustness rate (87.0\%) and third-highest standard accuracy (88.5\%, and with a <1\% drop compared to standard training). Our approach achieves the balance between accuracy and robustness. 
    }
}
\end{center}

\paragraph{RQ2:  How effective is our method in defending adversarial attacks?}

\begin{table}[t]
\centering
\caption{The defence success rate against AutoAttack~\cite{croce2020reliable} of different approaches on various datasets.}
\footnotesize
\begin{tabular}{l|cccc@{}}
\toprule
Approach &\multicolumn{4}{c}{Defence Success Rate} \\
\cline{2-5}
~ & \scriptsize CIFAR-100 & \scriptsize CIFAR-10 & \scriptsize SVHN & \scriptsize MNIST \\ 
\midrule
        ERM & 0.01 & 0.00 & 2.72 & 0.01 \\ 
        DA& 0.03 & 0.00 & 2.08 & 5.23 \\ 
        PGDT & 31.48 & 40.90 & 44.89 & 94.65 \\ 
        TRADES & 33.05 & 44.35 & 54.89 & 94.76 \\ 
        MART & 32.43 & 38.10 & 45.23 & 94.13 \\ 
        RS & 9.25 & 0.00 & 56.29 & 87.15 \\ 
        IBP & 29.33 & 37.10 & 61.94 & 89.18 \\ 
        PRL & 0.00 & 0.71 & 3.11 & 26.03 \\ 
        Ours & \textbf{53.05} & \textbf{88.08} & \textbf{92.15} & \textbf{97.8} \\ 
\midrule
\rowcolor{mygray}\multicolumn{5}{l}{$L^\infty$ bound at 0.3 for MNIST, and 8/255 for CIFAR or SVHN. }\\
\bottomrule
\end{tabular}
\label{tab:defence}
\end{table}
\begin{table*}[t]
\caption{Model defence success rates against adversarial attacks on standard benchmarks. Experiments are run on CIFAR-10 for all baselines and attacks.}
\label{tab:adaptive}
\centering
\footnotesize
\begin{tabular}{@{}l|ccccccccc}
\toprule
Attack & ERM & DA& PGDT & TRADES & MART & RS & IBP  & PRL & Ours  \\ 
\midrule
        No Attack & \textbf{94.85} & 94.21 & 84.38 & 80.42 & 81.54 & 89.45 & 48.40 & 93.82 & 94.23  \\ 
        TIFGSM~\cite{dong2019evading} & 35.10 & 33.00 & 65.70 & 62.90 & 69.10 & 45.40 & 40.20 & 34.00 & \textbf{92.80}  \\ 
        MIFGSM~\cite{dong2018boosting} & 0.00 & 0.00 & 50.90 & 51.90 & 50.50 & 5.80 & 38.10 & 0.00 & \textbf{92.80}  \\ 
        DIFGSM~\cite{xie2019improving} & 1.00 & 0.00 & 51.75 & 50.50 & 53.60 & 4.10 & 38.10 & 3.10 & \textbf{92.80}  \\ 
        VMIFGSM~\cite{wang2021enhancing} & 0.00 & 0.00 & 51.10 & 50.90 & 51.90 & 4.10 & 38.10 & 0.00 & \textbf{93.90}  \\ 
        TPGD & 38.10 & 39.20 & 69.30 & 69.10 & 70.10 & 48.50 & 50.00 & 28.90 & \textbf{91.80}  \\ 
        FGSM~\cite{goodfellow2014explaining} & 29.90 & 25.80 & 57.95 & 54.60 & 61.90 & 28.90 & 38.10 & 25.80 & \textbf{93.80}  \\ 
        RFGSM~\cite{tramer2017ensemble} & 0.00 & 0.00 & 49.15 & 50.40 & 48.50 & 3.70 & 38.10 & 0.00 & \textbf{90.00}  \\ 
        BIM~\cite{kurakin2016adversarial} & 0.00 & 0.00 & 52.00 & 57.20 & 47.40 & 2.10 & 38.10 & 0.00 & \textbf{90.70}  \\ 
        FAB~\cite{croce2020minimally} & 1.00 & 2.10 & 43.00 & 46.40 & 40.20 & 5.30 & 38.10 & 4.10 & \textbf{90.10}  \\ 
        CW~\cite{carlini2017towards} & 0.00 & 0.00 & 32.20 & 35.10 & 29.90 & 1.00 & 40.20 & 1.00 & \textbf{92.90}  \\ 
        UPGD & 0.00 & 0.00 & 49.85 & 50.50 & 49.80 & 5.10 & 38.10 & 0.00 & \textbf{93.80}  \\ 
        FFGSM~\cite{wong2020fast} & 19.60 & 23.70 & 60.55 & 55.70 & 66.00 & 33.00 & 42.30 & 29.90 & \textbf{92.80}  \\ 
        Jitter~\cite{schwinn2023exploring} & 11.30 & 12.40 & 48.15 & 47.40 & 49.50 & 34.00 & 39.20 & 24.70 & \textbf{90.70}  \\ 
        PGD & 0.00 & 0.00 & 57.40 & 54.60 & 60.80 & 7.20 & 40.20 & 0.00 & \textbf{91.80}  \\ 
        EOTPGD~\cite{liu2018adv} & 0.00 & 0.00 & 50.10 & 50.30 & 50.50 & 3.00 & 38.10 & 0.00 & \textbf{90.70}  \\ 
        APGD~\cite{croce2020reliable} & 0.00 & 0.00 & 48.40 & 51.00 & 46.40 & 1.00 & 38.10 & 0.00 & \textbf{90.70}  \\ 
        NIFGSM~\cite{lin2019nesterov} & 0.00 & 0.00 & 57.95 & 56.70 & 59.80 & 7.20 & 38.10 & 1.00 & \textbf{92.80}  \\ 
        SiniFGSM~\cite{lin2019nesterov} & 4.10 & 1.00 & 59.00 & 56.70 & 61.90 & 23.70 & 38.10 & 12.40 & \textbf{93.70}  \\ 
        VNIFGSM~\cite{wang2021enhancing} & 0.00 & 0.00 & 50.45 & 53.00 & 48.50 & 5.10 & 38.10 & 0.00 & \textbf{92.90}  \\ 
        APGDT~\cite{croce2020reliable} & 0.00 & 0.00 & 40.90 & 44.30 & 38.10 & 0.00 & 38.10 & 0.00 & \textbf{88.70}  \\ 
        Square~\cite{andriushchenko2020square} & 0.00 & 1.00 & 50.40 & 54.00 & 47.40 & 3.10 & 38.10 & 2.10 & \textbf{88.08}  \\ 
        Add Gaussian Noise & 25.80 & 43.30 & 79.10 & 78.40 & 80.40 & 74.20 & 42.30 & 45.40 & \textbf{87.60}  \\ 
        OnePixel~\cite{su2019one} & 79.40 & 83.50 & 78.05 & 74.20 & 82.50 & 83.50 & 42.50 & 80.40 & \textbf{89.70 } \\ 
        Pixle~\cite{pomponi2022pixle} & 0.00 & 0.00 & 12.55 & 11.30 & 14.40 & 1.00 & 10.30 & 0.00 & \textbf{17.50}  \\ 
        PGDL2 & 1.00 & 0.00 & 35.80 & 36.10 & 36.10 & 5.20 & 36.10 & 0.00 & \textbf{92.90}  \\ 
\midrule
\rowcolor{mygray}\multicolumn{10}{l}{
\begin{tabular}[l]{@{}c@{}}$L^\infty$ bound at 8/255. $L^2$ bound at 10/255. For Gaussian noise, std=0.1. More detailed parameter setting is according to \\ \url{https://adversarial-attacks-pytorch.readthedocs.io/en/latest/index.html}\end{tabular}

}\\
\bottomrule
\end{tabular}

\end{table*}
Two methods may achieve the same probabilistic robustness but have different behaviors when facing adversarial attacks, i.e., one may be easier to attack as those few adversarial samples may be easier to identify by existing adversarial attacking methods. We thus adopt the state-of-the-art method, AutoAttack~\cite{croce2020reliable}, to evaluate the effectiveness of our method and baselines in terms of defending against adversarial attacks. AutoAttack is an ensemble of different PGD attacks that is parameter-free, computationally affordable, and user-independent, making it an effective tool to assess adversarial robustness. The results, as presented in Table \ref{tab:defence}, demonstrate that our proposed method consistently achieves a higher defence success rate than any other approach across all four datasets. Specifically, our method achieves an impressive defence success rate of 82.77\% on average, surpassing the best performance of adversarial training, i.e., TRADES, which only attains a rate of 56.76\%. Apart from adversarial training, IBP achieves the highest defence success rate at in average 54.39\%. It is worth noting that although PRL exhibits relatively high certified robust accuracy for both $L^p$ and non-$L^p$ perturbation, second only to our method, their resilience against adversarial attacks is significantly low, with a defence success rate close to 0 on CIFAR-10 and CIFAR-100.

Additionally, we evaluate our proposed method under 25 adversarial attacks on CIFAR-10 and compare its defence success rates with several baseline methods, as shown in \cref{tab:adaptive}. Row No Attack is the standard accuracy on the original testing set. It is evident that our approach outperforms all baseline methods across all adversarial attack algorithms. Except for the Pixle~\cite{pomponi2022pixle} attack, our method achieves a defence success rate of over 88\% for all other attack methods. This is because Pixle attack focuses on searching for adversarial examples using the $L^0$-norm, which is not the focus of our method. Moreover, baseline methods with better average defence success rates, i.e., PGDT, TRADES, and MART, exhibit a significant decrease in standard accuracy (more than 10\%). PRL continues to show poor performance against these adversarial attacks, achieving a success rate of less than 5\% in most of the cases (17/25). This is because the adversarial examples of a PRL model, although only account for a small number (< 9.38\% on CIFAR-10, SVHN, or MNIST), are relatively easy to be searched by attack algorithms.

Overall, our method demonstrates robustness and effectiveness in defending against a wide range of adversarial attacks.
\begin{center}
\fcolorbox{black}{white!20}{\parbox{0.97\linewidth}
    {
        \emph{\textbf{Answer to RQ2}}:
        The proposed approach achieves the highest defence success rate (82.3\%). Its high certified robustness indeed brings benefits in defending against various adversarial attacks including AutoAttack~\cite{croce2020reliable}.
    }
}
\end{center}

\paragraph{RQ3: How efficient is our approach?}

\begin{figure*}[t]
  \centering
  \begin{subfigure}[b]{0.3\linewidth}
    \includegraphics[width=\linewidth]{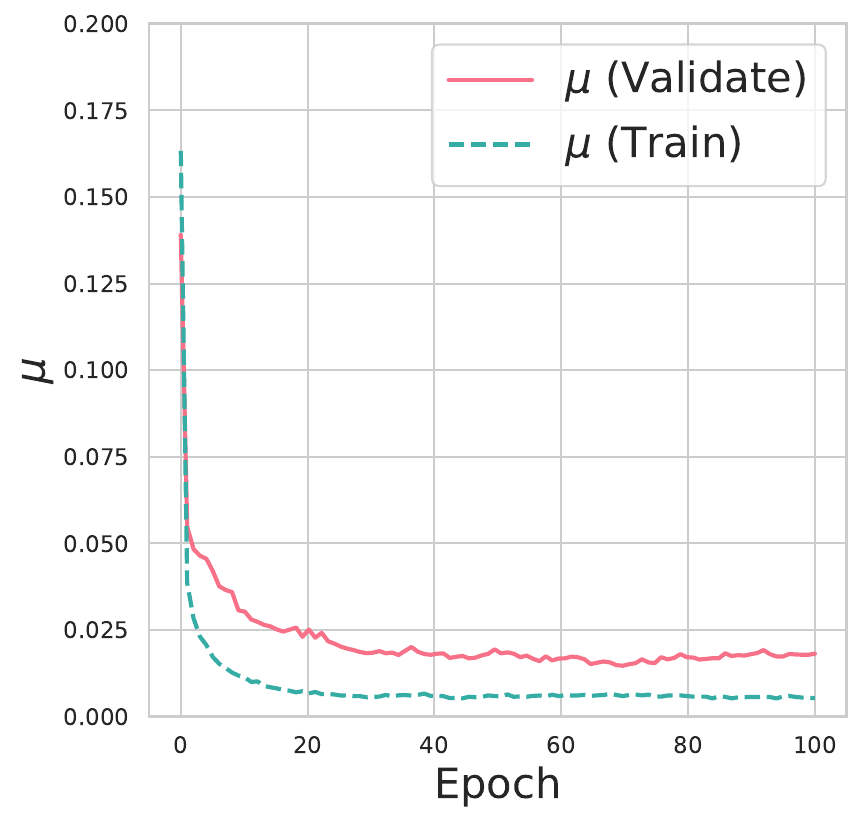}
    \caption{}
  \end{subfigure}
  \begin{subfigure}[b]{0.3\linewidth}
    \includegraphics[width=\linewidth]{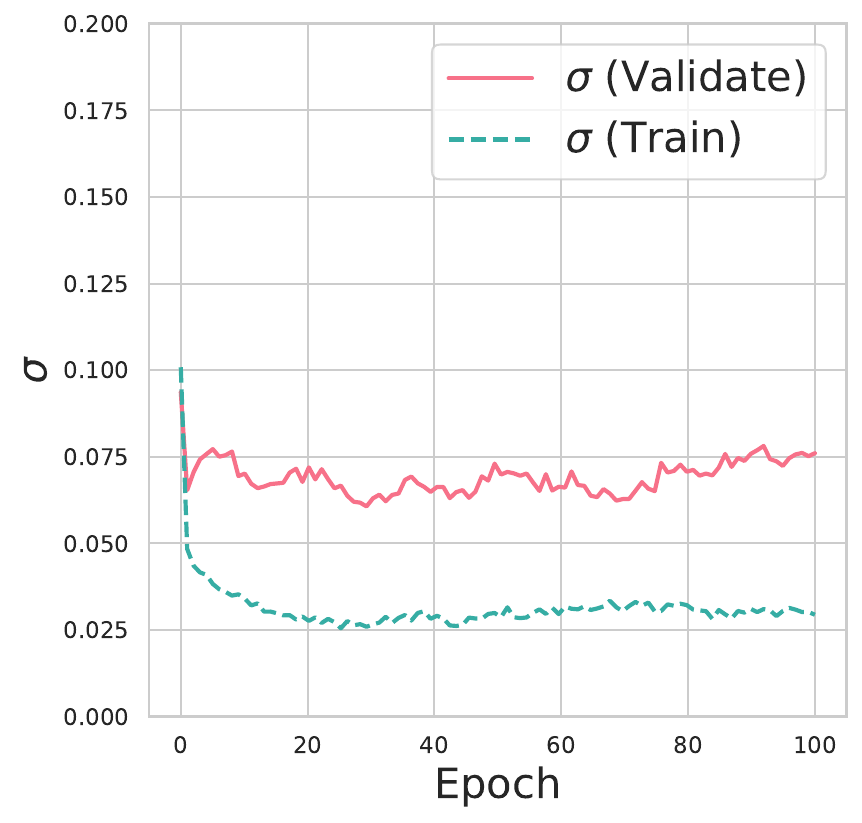}
    \caption{}
  \end{subfigure}
  \begin{subfigure}[b]{0.3\linewidth}
    \includegraphics[width=0.97\linewidth]{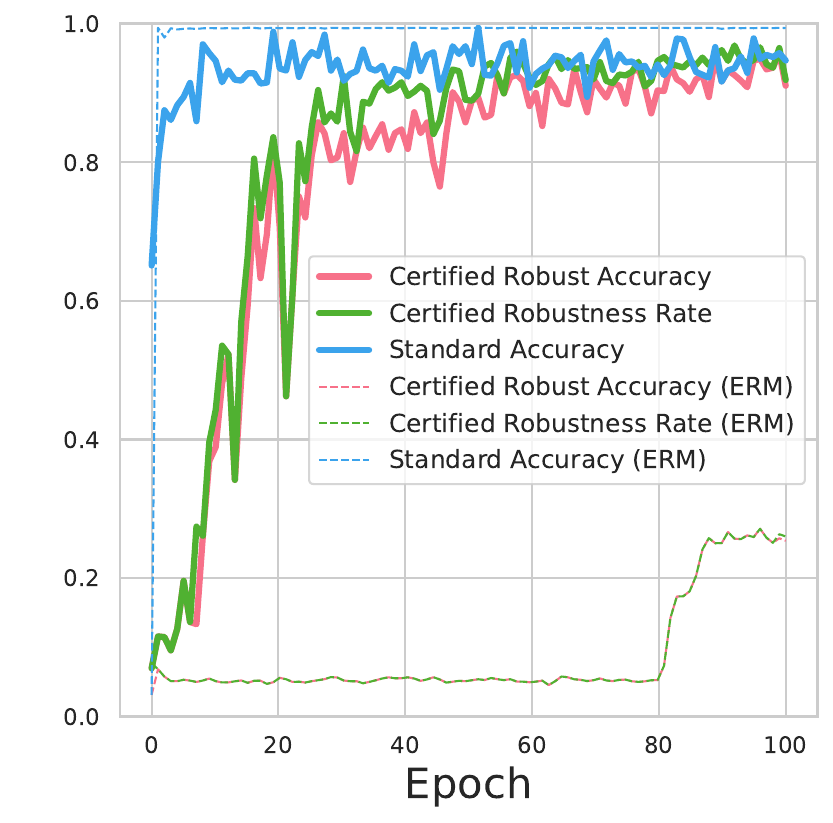}
    \caption{}
  \end{subfigure}
  \caption{(a) \& (b) Illustrations of loss convergence of the mean term and variance term in \cref{objective}. (c) Illustration of model performance in terms of certified robust accuracy, certified robustness rate as well as standard accuracy. The illustrated figures are from experiments on MNIST. The performance of ERM~\cite{vapnik1999nature} is also shown for comparison.}
  \label{fig:converge}
\end{figure*}

\begin{table}[t]
    \centering
    \caption{Comparison of overhead for our training on MNIST dataset with 300 epoch training time. For a fair comparison, the training cost of all approaches is collected using a single NVIDIA RTX 2080 Ti GPU. }
    \footnotesize
    \begin{tabular}{l|c|c}
    \toprule    
        Approach & Training time (sec) & Inference time (sec) \\ 
        \midrule
        ERM & $4.9\times10^2$ & 27 \\ 
        DA& $8.9\times10^3$ & 27 \\ 
        PGD & $9.9\times10^3$ & 27 \\ 
        TRADES & $9.9\times10^3$ & 27 \\ 
        MART & $9.9\times10^3$ & 27 \\ 
        RS & $5.8\times10^4$ & $2.7\times10^5$ \\ 
        IBP & $2.1\times10^5$ & 27 \\ 
        PRL & $2.8\times10^4$ & $2.7\times10^3$ \\ 
        Ours & $9.6\times10^3$ & $4.7\times10^3$   \\ \bottomrule
    \end{tabular}
    \label{tab:overhead}
\end{table}

To answer this question, we measure the training and inference time of our method and the baseline approaches on MNIST. For inference, the time is collected on the whole testing set comprised of 10,000 samples. The results are shown in \cref{tab:overhead}. 

\noindent \textbf{Training time.} For training efficiency, we can observe that compared to methods designed to certify robustness, i.e., IBP and PRL, our method demonstrates significantly higher training efficiency, being 21.93 and 2.89 times faster, respectively. 
Our method has a similar training cost to data augmentation and adversarial training, with a total training time around 10 thousand seconds. 
These findings indicate that our approach is highly efficient and practical for training deep neural networks with robustness guarantees.

\noindent \textbf{Inference time.} It is worth noting that the inherent inference process of ERM, DA, and adversarial training algorithms does not provide robustness certification. We thus focus on comparing our method with certified training algorithms, i.e., RS, IBP, and PRL. From \cref{tab:overhead}, it can be observed that IBP takes the least inference time as it only requires a single forward propagation on the input to obtain the predictions and certification results. In contrast, the other three methods provide certification by predicting a large number of samples around the input. This efficiency is achieved by trading off training time. Our inference can be considered reasonably efficient, as it has the same order of magnitude as PRL and is two orders of magnitude faster than RS. This is mainly attributed to our method using sequential sampling to reduce processing time. 
That is, sequential sampling allows for decisions to be made based on observed data at each step~\cite{wald1992sequential} and is known to reduce required sample sizes while maintaining statistical correctness due to its adaptability~\cite{mead1990design}. In comparison to fixed-size sampling, sequential sampling may lead to increased efficiency~\cite{chernoff1959sequential}.

Our method converges efficiently on correctly predicting unperturbed variables, but convergence on perturbed variables is slightly delayed, as illustrated in \cref{fig:converge}(c).
\begin{center}
\fcolorbox{black}{white!20}{\parbox{0.97\linewidth}
    {
        \emph{\textbf{Answer to RQ3}}:
        Our approach has a training cost similar to data augmentation and adversarial training, and is much more efficient than certified training, making it practical for training neural networks with robustness guarantees.
    }
}
\end{center}

\paragraph{RQ4: How do the hyper-parameters impact the performance of our approach?}

\begin{table}[t]
\centering
\caption{The certified robust accuracy and certified robustness rate of different approaches on various datasets within smaller vicinity.}
\footnotesize
\begin{tabular}{l| cccc}
\toprule
Approach &\multicolumn{4}{c}{Certified Robust Accuracy}\\
\midrule
~ & \scriptsize CIFAR-100 & \scriptsize CIFAR-10 & \scriptsize SVHN & \scriptsize MNIST \\ 
        ERM & 33.45 & 48.85 & 59.34 & 48.01 \\
        DA& 54.43 & 83.50 & 84.79 & 81.23 \\
        PGDT & 44.59 & 83.23 & 87.98 & 95.89 \\
        TRADES & 58.86 & 80.57 & 82.45 & 95.39 \\
        MART & 56.73 & 81.35 & 73.84 & 95.22 \\
        RS & 53.93 & 88.98 & 86.03 & 90.48 \\
        IBP & 33.45 & 54.41 & 67.34 & 97.74 \\
        PRL & 53.99 & 91.74 & 91.97 & 98.99 \\
        Ours & \textbf{57.27} & \textbf{93.58} & \textbf{92.85} & \textbf{97.15} \\
\midrule
\rowcolor{mygray}\multicolumn{5}{l}{\begin{tabular}[c]{@{}l@{}}$\kappa = 10^{-2}, 1 - \alpha = 0.99$; $L^\infty$ bound at 0.1 for MNIST, and 2/255 \\ for CIFAR and SVHN. See \cref{tab:certify} for $L^\infty$ bound at 0.3 for MNIST,\\ and 8/255 for CIFAR and SVHN. \end{tabular} }\\
\bottomrule
\end{tabular}
\label{tab:vicinity}
\end{table}

\begin{figure}
    \centering
    \includegraphics[width=0.6\linewidth]{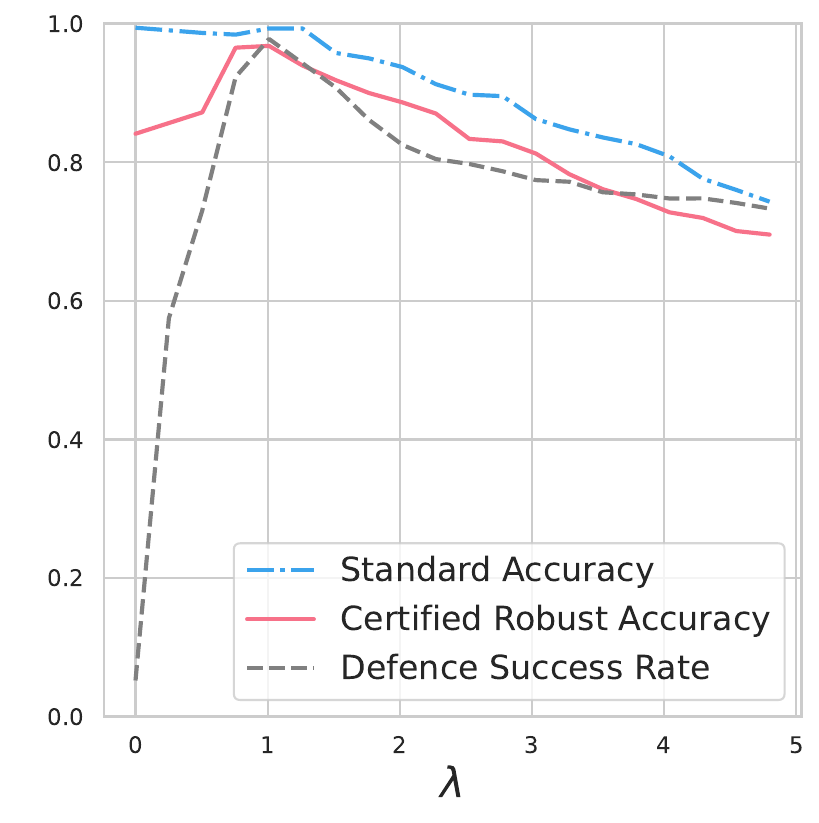}
    \caption{Adjusting hyperparameter $\lambda$ makes the model converge with different performance. The experiment is run on MNIST with an $L^\infty$ bound of 0.3.}
    \label{fig:lambda}
\end{figure}

We carry out an ablation study to assess the effect of the hyper-parameters in our method. \\

\noindent \textbf{Vicinity size $\epsilon$.} To investigate the impact of the vicinity size on certified robust accuracy, we evaluate the models with altered $L^\infty$-norm radius $\epsilon$ on each dataset. Specifically, for MNIST, values of $\epsilon$ are selected from $\set{0.1, 0.3}$, while for the other three datasets, its values are chosen from $\set{2/255, 8/255}$. The results are shown in \cref{tab:vicinity}. We observe a trade-off between certified robust accuracy and the usefulness of certification, i.e., decreasing the vicinity radius increases certified robust accuracy. 
Our approach achieves high certified robust accuracy ($>85\%$) within a reasonable range of the vicinity and experiences a $0.36\%$ increase with a one-third reduction and a $2.98\%$ average increase with a one-quarter reduction.

\noindent \textbf{Importance factor $\lambda$.} We conduct an experiment on the impact of importance factor $\lambda$ on the performance of neural models, including standard accuracy, certified robust accuracy, and defence success rate against AutoAttack. The value of $\lambda$ ranges from 0 to 5, with an interval of 0.25. \cref{fig:lambda} presents the trend of changes in model performance for MNIST, which is representative of other results. Note that a value of $\lambda$ close to 1 yields the best performance. When the value of $\lambda$ decreases, the contribution of the proposed variance-minimization term decreases as well. If $\lambda$ is too small, i.e., close to 0, the training process becomes similar to data-augmented training with random perturbation, which prioritizes optimizing average losses, resulting in a drop in the certified robust rate and thus decreasing the certified robustness accuracy. On the other hand, if the loss function excessively emphasizes the variance term with a large value of $\lambda$, it can lead to a decrease in standard accuracy and further impact the certified robust accuracy. Additionally, the defence success rate also decreases by about a quarter when varying $\lambda$ from 1 to 5.

\noindent \textbf{Percentage to certify $\kappa$.} To investigate how the strictness of certification requirement influences the certified robust accuracy, we vary the acceptable level $\kappa$ and significance level $\alpha$. The certified robust accuracy with regard to different acceptable level and significance level is presented in \cref{tab:strictness} and \cref{tab:alpha}, respectively. Note that $\kappa=0$ means conducting deterministic robustness certification on the model, which can only be achieved by IBP. The remaining baselines and our method can only provide probabilistic robustness certification results for the model. It can be observed that the variation of both the acceptable level $\kappa$ and significance level $\alpha$ does not have a significant impact on the certified robust accuracy, except ERM and DA. Specifically, for our method, when $\kappa$ has changed from $10^{-3}$ to $10^{-1}$, the certified robust accuracy has only improved by 1.05\%; no increase in certified robust accuracy is observed when $\alpha$ varies from $10^{-3}$ to $5\times10^{-2}$.

\begin{table}[t]
    \centering
    \caption{Comparison of the influence of different $\kappa$ values on the certified robust accuracy of CIFAR-10. }
    \footnotesize
    \begin{tabular}{l|c|c|c|c@{}}
    \toprule    
         Approach & \begin{tabular}[c]{@{}l@{}}$\kappa = 0$\\ \scriptsize(Deterministic)\end{tabular} & \begin{tabular}[c]{@{}l@{}}$\kappa = 10^{-3}$\\~\end{tabular} & \begin{tabular}[c]{@{}l@{}}$\kappa = 10^{-2}$\\~\end{tabular} & \begin{tabular}[c]{@{}l@{}}$\kappa = 10^{-1}$\\~\end{tabular} \\ \midrule
        ERM & - & 1.25 & 1.25 & 25.09 \\ 
        DA& - & 73.50 & 76.07 & 86.59 \\ 
        PGDT & - & 82.82 & 82.90 & 82.95 \\ 
        TRADES & - & 78.69 & 78.80 & 79.60 \\ 
        MART & - & 71.42 & 72.21 & 73.43 \\ 
        RS & - & 87.63 & 87.98 & 88.08 \\ 
        IBP & 35.13 & 39.98 & 40.00 & 44.41 \\ 
        PRL & - & 89.88 & 90.63 & 91.97 \\ 
        Ours & - & 91.73 & 91.75 & 92.78 \\  \midrule
        \multicolumn{5}{l}{\cellcolor{mygray}For $\kappa > 0$, $\alpha$ takes $10^{-2}$.}\\\bottomrule
    \end{tabular}
    \label{tab:strictness}
\end{table}

\begin{table}[t]
\centering
\caption{Comparison of the influence of different $\alpha$ values on the certified robust accuracy of CIFAR-10 where $\kappa = 10^{-2}$. }
\footnotesize
\begin{tabular}{l|c|c|c@{}}
\toprule
Approach & $1-\alpha = 0.95$ & $1-\alpha = 0.99$ & $1-\alpha = 0.999$\\ \midrule
ERM & 2.55 & 1.25 & 1.25 \\
DA& 77.56 & 76.07 & 76.07 \\
PGDT & 82.90 & 82.90 & 82.90 \\
TRADES & 78.80 & 78.80 & 78.80 \\
MART & 72.21 & 72.21 & 72.21 \\
RS & 87.98 & 87.98 & 87.98 \\
IBP & 40.00 & 40.00 & 40.00 \\
PRL & 90.63 & 90.63 & 90.63 \\
Ours & 91.75 & 91.75 & 91.75 \\ \bottomrule
\end{tabular}
\label{tab:alpha}
\end{table}

\section{Conclusion}
We present an approach that improves the robustness of neural networks against adversarial examples. Our approach includes a training method that minimizes both the mean and variance of the loss in prediction and an inference method that provides probabilistic-certified robustness. Through theoretical analysis, we have shown that minimizing variance is the upper bound of the probability of adversarial examples and that higher quantile accuracy leads to over 91\% certified robust accuracy. Our experimental results on standard benchmark datasets show that our method achieves higher defence success rate and certification rate compared to the state-of-the-art while sacrificing less standard accuracy.

\bibliographystyle{plain}
\bibliography{favourite,custom}

\begin{thebibliography}{10}

\bibitem{andriushchenko2020square}
Maksym Andriushchenko, Francesco Croce, Nicolas Flammarion, and Matthias Hein.
\newblock Square attack: A query-efficient black-box adversarial attack via
  random search.
\newblock In Andrea Vedaldi, Horst Bischof, Thomas Brox, and Jan-Michael Frahm,
  editors, {\em Computer Vision -- ECCV 2020}, pages 484--501, Cham, 2020.
  Springer International Publishing.

\bibitem{athalye2018obfuscated}
Anish Athalye, Nicholas Carlini, and David Wagner.
\newblock Obfuscated gradients give a false sense of security: Circumventing
  defenses to adversarial examples.
\newblock In Jennifer Dy and Andreas Krause, editors, {\em Proceedings of the
  35th International Conference on Machine Learning}, volume~80 of {\em
  Proceedings of Machine Learning Research}, pages 274--283. PMLR, 10--15 Jul
  2018.

\bibitem{athalye2018synthesizing}
Anish Athalye, Logan Engstrom, Andrew Ilyas, and Kevin Kwok.
\newblock Synthesizing robust adversarial examples.
\newblock In Jennifer Dy and Andreas Krause, editors, {\em Proceedings of the
  35th International Conference on Machine Learning}, volume~80 of {\em
  Proceedings of Machine Learning Research}, pages 284--293. PMLR, 10--15 Jul
  2018.

\bibitem{bai2021recent}
Tao Bai, Jinqi Luo, Jun Zhao, Bihan Wen, and Qian Wang.
\newblock Recent advances in adversarial training for adversarial robustness.
\newblock In Zhi-Hua Zhou, editor, {\em Proceedings of the Thirtieth
  International Joint Conference on Artificial Intelligence, {IJCAI-21}}, pages
  4312--4321. International Joint Conferences on Artificial Intelligence
  Organization, 8 2021.
\newblock Survey Track.

\bibitem{balunovic2020adversarial}
Mislav Balunovic and Martin~T. Vechev.
\newblock Adversarial training and provable defenses: Bridging the gap.
\newblock In {\em 8th International Conference on Learning Representations,
  {ICLR} 2020, Addis Ababa, Ethiopia, April 26-30, 2020}. OpenReview.net, 2020.

\bibitem{bhattacharya2019survey}
Saswati Bhattacharya and Mousumi Gupta.
\newblock A survey on: Facial emotion recognition invariant to pose,
  illumination and age.
\newblock In {\em 2019 Second International Conference on Advanced
  Computational and Communication Paradigms (ICACCP)}, pages 1--6. IEEE, 2019.

\bibitem{blitzstein2019introduction}
Joseph~K Blitzstein and Jessica Hwang.
\newblock {\em Introduction to probability}.
\newblock Crc Press, 2019.

\bibitem{carlini2019evaluating}
Nicholas Carlini, Anish Athalye, Nicolas Papernot, Wieland Brendel, Jonas
  Rauber, Dimitris Tsipras, Ian~J. Goodfellow, Aleksander Madry, and Alexey
  Kurakin.
\newblock On evaluating adversarial robustness.
\newblock {\em CoRR}, abs/1902.06705, 2019.

\bibitem{carlini2017towards}
Nicholas Carlini and David Wagner.
\newblock Towards evaluating the robustness of neural networks.
\newblock In {\em 2017 IEEE Symposium on Security and Privacy (SP)}, pages
  39--57, 2017.

\bibitem{chernoff1959sequential}
Herman Chernoff.
\newblock Sequential design of experiments.
\newblock {\em The Annals of Mathematical Statistics}, 30(3):755--770, 1959.

\bibitem{chiang2020certified}
Ping{-}yeh Chiang, Renkun Ni, Ahmed Abdelkader, Chen Zhu, Christoph Studer, and
  Tom Goldstein.
\newblock Certified defenses for adversarial patches.
\newblock In {\em 8th International Conference on Learning Representations,
  {ICLR} 2020, Addis Ababa, Ethiopia, April 26-30, 2020}. OpenReview.net, 2020.

\bibitem{cohen2019certified}
Jeremy Cohen, Elan Rosenfeld, and Zico Kolter.
\newblock Certified adversarial robustness via randomized smoothing.
\newblock In Kamalika Chaudhuri and Ruslan Salakhutdinov, editors, {\em
  Proceedings of the 36th International Conference on Machine Learning},
  volume~97 of {\em Proceedings of Machine Learning Research}, pages
  1310--1320. PMLR, 09--15 Jun 2019.

\bibitem{croce2020minimally}
Francesco Croce and Matthias Hein.
\newblock Minimally distorted adversarial examples with a fast adaptive
  boundary attack.
\newblock In Hal~Daumé III and Aarti Singh, editors, {\em Proceedings of the
  37th International Conference on Machine Learning}, volume 119 of {\em
  Proceedings of Machine Learning Research}, pages 2196--2205. PMLR, 13--18 Jul
  2020.

\bibitem{croce2020reliable}
Francesco Croce and Matthias Hein.
\newblock Reliable evaluation of adversarial robustness with an ensemble of
  diverse parameter-free attacks.
\newblock In Hal~Daumé III and Aarti Singh, editors, {\em Proceedings of the
  37th International Conference on Machine Learning}, volume 119 of {\em
  Proceedings of Machine Learning Research}, pages 2206--2216. PMLR, 13--18 Jul
  2020.

\bibitem{dong2018boosting}
Yinpeng Dong, Fangzhou Liao, Tianyu Pang, Hang Su, Jun Zhu, Xiaolin Hu, and
  Jianguo Li.
\newblock Boosting adversarial attacks with momentum.
\newblock In {\em Proceedings of the IEEE Conference on Computer Vision and
  Pattern Recognition (CVPR)}, June 2018.

\bibitem{dong2019evading}
Yinpeng Dong, Tianyu Pang, Hang Su, and Jun Zhu.
\newblock Evading defenses to transferable adversarial examples by
  translation-invariant attacks.
\newblock In {\em Proceedings of the IEEE/CVF Conference on Computer Vision and
  Pattern Recognition (CVPR)}, June 2019.

\bibitem{evtimov2018robust}
Ivan Evtimov, Kevin Eykholt, Earlence Fernandes, Tadayoshi Kohno, Bo~Li, Atul
  Prakash, Amir Rahmati, and Dawn Song.
\newblock Robust physical-world attacks on deep learning visual classification.
\newblock In {\em Proceedings of the IEEE Conference on Computer Vision and
  Pattern Recognition}, pages 1625--1634, 2018.

\bibitem{ganin2016domain}
Yaroslav Ganin, Evgeniya Ustinova, Hana Ajakan, Pascal Germain, Hugo
  Larochelle, Fran{\c{c}}ois Laviolette, Mario March, and Victor Lempitsky.
\newblock Domain-adversarial training of neural networks.
\newblock {\em Journal of Machine Learning Research}, 17(59):1--35, 2016.

\bibitem{goodfellow2014explaining}
Ian~J. Goodfellow, Jonathon Shlens, and Christian Szegedy.
\newblock Explaining and harnessing adversarial examples.
\newblock In Yoshua Bengio and Yann LeCun, editors, {\em 3rd International
  Conference on Learning Representations, {ICLR} 2015, San Diego, CA, USA, May
  7-9, 2015, Conference Track Proceedings}, 2015.

\bibitem{he2016deep}
Kaiming He, Xiangyu Zhang, Shaoqing Ren, and Jian Sun.
\newblock Deep residual learning for image recognition.
\newblock In {\em Proceedings of the IEEE Conference on Computer Vision and
  Pattern Recognition (CVPR)}, June 2016.

\bibitem{krizhevsky2009learning}
Alex Krizhevsky, Geoffrey Hinton, et~al.
\newblock Learning multiple layers of features from tiny images.
\newblock 2009.

\bibitem{kurakin2018adversarial}
Alexey Kurakin, Ian~J. Goodfellow, and Samy Bengio.
\newblock Adversarial examples in the physical world.
\newblock In {\em 5th International Conference on Learning Representations,
  {ICLR} 2017, Toulon, France, April 24-26, 2017, Workshop Track Proceedings}.
  OpenReview.net, 2017.

\bibitem{kurakin2016adversarial}
Alexey Kurakin, Ian~J. Goodfellow, and Samy Bengio.
\newblock Adversarial machine learning at scale.
\newblock In {\em 5th International Conference on Learning Representations,
  {ICLR} 2017, Toulon, France, April 24-26, 2017, Conference Track
  Proceedings}. OpenReview.net, 2017.

\bibitem{lecunmnist}
Yann LeCun, Corinna Cortes, and Chris Burges.

\bibitem{li2023sok}
Linyi Li, Tao Xie, and Bo~Li.
\newblock Sok: Certified robustness for deep neural networks.
\newblock In {\em 44th {IEEE} Symposium on Security and Privacy, {SP} 2023, San
  Francisco, CA, USA, 22-26 May 2023}. IEEE, 2023.

\bibitem{lin2019nesterov}
Jiadong Lin, Chuanbiao Song, Kun He, Liwei Wang, and John~E. Hopcroft.
\newblock Nesterov accelerated gradient and scale invariance for adversarial
  attacks.
\newblock In {\em International Conference on Learning Representations}, 2020.

\bibitem{liu2019adaptiveface}
Hao Liu, Xiangyu Zhu, Zhen Lei, and Stan~Z. Li.
\newblock Adaptiveface: Adaptive margin and sampling for face recognition.
\newblock In {\em Proceedings of the IEEE/CVF Conference on Computer Vision and
  Pattern Recognition (CVPR)}, June 2019.

\bibitem{liu2018adv}
Xuanqing Liu, Yao Li, Chongruo Wu, and Cho{-}Jui Hsieh.
\newblock Adv-bnn: Improved adversarial defense through robust bayesian neural
  network.
\newblock In {\em 7th International Conference on Learning Representations,
  {ICLR} 2019, New Orleans, LA, USA, May 6-9, 2019}. OpenReview.net, 2019.

\bibitem{ma2018characterizing}
Xingjun Ma, Bo~Li, Yisen Wang, Sarah~M. Erfani, Sudanthi N.~R. Wijewickrema,
  Grant Schoenebeck, Dawn Song, Michael~E. Houle, and James Bailey.
\newblock Characterizing adversarial subspaces using local intrinsic
  dimensionality.
\newblock In {\em 6th International Conference on Learning Representations,
  {ICLR} 2018, Vancouver, BC, Canada, April 30 - May 3, 2018, Conference Track
  Proceedings}. OpenReview.net, 2018.

\bibitem{madry2017towards}
Aleksander Madry, Aleksandar Makelov, Ludwig Schmidt, Dimitris Tsipras, and
  Adrian Vladu.
\newblock Towards deep learning models resistant to adversarial attacks.
\newblock In {\em 6th International Conference on Learning Representations,
  {ICLR} 2018, Vancouver, BC, Canada, April 30 - May 3, 2018, Conference Track
  Proceedings}. OpenReview.net, 2018.

\bibitem{mead1990design}
Roger Mead.
\newblock {\em The design of experiments: statistical principles for practical
  applications}.
\newblock Cambridge university press, 1990.

\bibitem{miyato2018virtual}
Takeru Miyato, Shin-Ichi Maeda, Masanori Koyama, and Shin Ishii.
\newblock Virtual adversarial training: A regularization method for supervised
  and semi-supervised learning.
\newblock {\em IEEE Transactions on Pattern Analysis and Machine Intelligence},
  41(8):1979--1993, 2019.

\bibitem{muller2022certified}
Mark~Niklas M{\"{u}}ller, Franziska Eckert, Marc Fischer, and Martin~T. Vechev.
\newblock Certified training: Small boxes are all you need.
\newblock {\em CoRR}, abs/2210.04871, 2022.

\bibitem{netzer2011reading}
Yuval Netzer, Tao Wang, Adam Coates, Alessandro Bissacco, Bo~Wu, and Andrew~Y.
  Ng.
\newblock Reading digits in natural images with unsupervised feature learning.
\newblock In {\em NIPS Workshop on Deep Learning and Unsupervised Feature
  Learning 2011}, 2011.

\bibitem{pang2022robustness}
Tianyu Pang, Min Lin, Xiao Yang, Jun Zhu, and Shuicheng Yan.
\newblock Robustness and accuracy could be reconcilable by (proper) definition.
\newblock In Kamalika Chaudhuri, Stefanie Jegelka, Le~Song, Csaba
  Szepesv{\'{a}}ri, Gang Niu, and Sivan Sabato, editors, {\em International
  Conference on Machine Learning, {ICML} 2022, 17-23 July 2022, Baltimore,
  Maryland, {USA}}, volume 162 of {\em Proceedings of Machine Learning
  Research}, pages 17258--17277. {PMLR}, 2022.

\bibitem{papernot2016limitations}
Nicolas Papernot, Patrick McDaniel, Somesh Jha, Matt Fredrikson, Z.~Berkay
  Celik, and Ananthram Swami.
\newblock The limitations of deep learning in adversarial settings.
\newblock In {\em 2016 IEEE European Symposium on Security and Privacy
  (EuroS\&P)}, pages 372--387. IEEE, 2016.

\bibitem{papernot2016transferability}
Nicolas Papernot, Patrick~D. McDaniel, and Ian~J. Goodfellow.
\newblock Transferability in machine learning: from phenomena to black-box
  attacks using adversarial samples.
\newblock {\em CoRR}, abs/1605.07277, 2016.

\bibitem{pomponi2022pixle}
Jary Pomponi, Simone Scardapane, and Aurelio Uncini.
\newblock Pixle: a fast and effective black-box attack based on rearranging
  pixels.
\newblock In {\em 2022 International Joint Conference on Neural Networks
  (IJCNN)}, pages 1--7, 2022.

\bibitem{raghunathan2018certified}
Aditi Raghunathan, Jacob Steinhardt, and Percy Liang.
\newblock Certified defenses against adversarial examples.
\newblock In {\em 6th International Conference on Learning Representations,
  {ICLR} 2018, Vancouver, BC, Canada, April 30 - May 3, 2018, Conference Track
  Proceedings}. OpenReview.net, 2018.

\bibitem{robey2022probabilistically}
Alexander Robey, Luiz Chamon, George~J. Pappas, and Hamed Hassani.
\newblock Probabilistically robust learning: Balancing average and worst-case
  performance.
\newblock In Kamalika Chaudhuri, Stefanie Jegelka, Le~Song, Csaba Szepesvari,
  Gang Niu, and Sivan Sabato, editors, {\em Proceedings of the 39th
  International Conference on Machine Learning}, volume 162 of {\em Proceedings
  of Machine Learning Research}, pages 18667--18686. PMLR, 17--23 Jul 2022.

\bibitem{schwinn2023exploring}
Leo Schwinn, Ren{\'e} Raab, An~Nguyen, Dario Zanca, and Bjoern Eskofier.
\newblock Exploring misclassifications of robust neural networks to enhance
  adversarial attacks.
\newblock {\em Applied Intelligence}, Mar 2023.

\bibitem{sharif2016accessorize}
Mahmood Sharif, Sruti Bhagavatula, Lujo Bauer, and Michael~K. Reiter.
\newblock Accessorize to a crime: Real and stealthy attacks on state-of-the-art
  face recognition.
\newblock In {\em Proceedings of the 2016 ACM SIGSAC Conference on Computer and
  Communications Security}, CCS '16, pages 1528--1540, New York, NY, USA, 2016.
  Association for Computing Machinery.

\bibitem{shi2021fast}
Zhouxing Shi, Yihan Wang, Huan Zhang, Jinfeng Yi, and Cho-Jui Hsieh.
\newblock Fast certified robust training with short warmup.
\newblock In M.~Ranzato, A.~Beygelzimer, Y.~Dauphin, P.S. Liang, and J.~Wortman
  Vaughan, editors, {\em Advances in Neural Information Processing Systems},
  volume~34, pages 18335--18349. Curran Associates, Inc., 2021.

\bibitem{shorten2019survey}
Connor Shorten and Taghi~M. Khoshgoftaar.
\newblock A survey on image data augmentation for deep learning.
\newblock {\em Journal of Big Data}, 6(1):60, Jul 2019.

\bibitem{silva2020opportunities}
Samuel~Henrique Silva and Peyman Najafirad.
\newblock Opportunities and challenges in deep learning adversarial robustness:
  {A} survey.
\newblock {\em CoRR}, abs/2007.00753, 2020.

\bibitem{singh2019abstract}
Gagandeep Singh, Timon Gehr, Markus P\"{u}schel, and Martin Vechev.
\newblock An abstract domain for certifying neural networks.
\newblock {\em Proc. ACM Program. Lang.}, 3(POPL), jan 2019.

\bibitem{su2019one}
Jiawei Su, Danilo~Vasconcellos Vargas, and Kouichi Sakurai.
\newblock One pixel attack for fooling deep neural networks.
\newblock {\em IEEE Transactions on Evolutionary Computation}, 23(5):828--841,
  2019.

\bibitem{szegedy2013intriguing}
Christian Szegedy, Wojciech Zaremba, Ilya Sutskever, Joan Bruna, Dumitru Erhan,
  Ian~J. Goodfellow, and Rob Fergus.
\newblock Intriguing properties of neural networks.
\newblock In Yoshua Bengio and Yann LeCun, editors, {\em 2nd International
  Conference on Learning Representations, {ICLR} 2014, Banff, AB, Canada, April
  14-16, 2014, Conference Track Proceedings}, 2014.

\bibitem{tramer2020adaptive}
Florian Tramer, Nicholas Carlini, Wieland Brendel, and Aleksander Madry.
\newblock On adaptive attacks to adversarial example defenses.
\newblock In H.~Larochelle, M.~Ranzato, R.~Hadsell, M.F. Balcan, and H.~Lin,
  editors, {\em Advances in Neural Information Processing Systems}, volume~33,
  pages 1633--1645. Curran Associates, Inc., 2020.

\bibitem{tramer2017ensemble}
Florian Tram{\`{e}}r, Alexey Kurakin, Nicolas Papernot, Ian~J. Goodfellow, Dan
  Boneh, and Patrick~D. McDaniel.
\newblock Ensemble adversarial training: Attacks and defenses.
\newblock In {\em 6th International Conference on Learning Representations,
  {ICLR} 2018, Vancouver, BC, Canada, April 30 - May 3, 2018, Conference Track
  Proceedings}. OpenReview.net, 2018.

\bibitem{tsipras2018robustness}
Dimitris Tsipras, Shibani Santurkar, Logan Engstrom, Alexander Turner, and
  Aleksander Madry.
\newblock Robustness may be at odds with accuracy.
\newblock In {\em 7th International Conference on Learning Representations,
  {ICLR} 2019, New Orleans, LA, USA, May 6-9, 2019}. OpenReview.net, 2019.

\bibitem{vaishnavi2022accelerating}
Pratik Vaishnavi, Kevin Eykholt, and Amir Rahmati.
\newblock Accelerating certified robustness training via knowledge transfer.
\newblock {\em CoRR}, abs/2210.14283, 2022.

\bibitem{vapnik1999nature}
Vladimir Vapnik.
\newblock {\em The nature of statistical learning theory}.
\newblock Springer science \& business media, 1999.

\bibitem{wald1992sequential}
A.~Wald.
\newblock {Sequential Tests of Statistical Hypotheses}.
\newblock {\em The Annals of Mathematical Statistics}, 16(2):117 -- 186, 1945.

\bibitem{wang2021enhancing}
Xiaosen Wang and Kun He.
\newblock Enhancing the transferability of adversarial attacks through variance
  tuning.
\newblock In {\em Proceedings of the IEEE/CVF Conference on Computer Vision and
  Pattern Recognition (CVPR)}, pages 1924--1933, June 2021.

\bibitem{wang2020improving}
Yisen Wang, Difan Zou, Jinfeng Yi, James Bailey, Xingjun Ma, and Quanquan Gu.
\newblock Improving adversarial robustness requires revisiting misclassified
  examples.
\newblock In {\em 8th International Conference on Learning Representations,
  {ICLR} 2020, Addis Ababa, Ethiopia, April 26-30, 2020}. OpenReview.net, 2020.

\bibitem{wen2020time}
Qingsong Wen, Liang Sun, Fan Yang, Xiaomin Song, Jingkun Gao, Xue Wang, and
  Huan Xu.
\newblock Time series data augmentation for deep learning: A survey.
\newblock In Zhi-Hua Zhou, editor, {\em Proceedings of the Thirtieth
  International Joint Conference on Artificial Intelligence, {IJCAI-21}}, pages
  4653--4660. International Joint Conferences on Artificial Intelligence
  Organization, 8 2021.
\newblock Survey Track.

\bibitem{wong2020fast}
Eric Wong, Leslie Rice, and J.~Zico Kolter.
\newblock Fast is better than free: Revisiting adversarial training.
\newblock In {\em 8th International Conference on Learning Representations,
  {ICLR} 2020, Addis Ababa, Ethiopia, April 26-30, 2020}. OpenReview.net, 2020.

\bibitem{xie2019improving}
Cihang Xie, Zhishuai Zhang, Yuyin Zhou, Song Bai, Jianyu Wang, Zhou Ren, and
  Alan~L. Yuille.
\newblock Improving transferability of adversarial examples with input
  diversity.
\newblock In {\em Proceedings of the IEEE/CVF Conference on Computer Vision and
  Pattern Recognition (CVPR)}, June 2019.

\bibitem{xu2020automatic}
Kaidi Xu, Zhouxing Shi, Huan Zhang, Yihan Wang, Kai-Wei Chang, Minlie Huang,
  Bhavya Kailkhura, Xue Lin, and Cho-Jui Hsieh.
\newblock Automatic perturbation analysis for scalable certified robustness and
  beyond.
\newblock In H.~Larochelle, M.~Ranzato, R.~Hadsell, M.F. Balcan, and H.~Lin,
  editors, {\em Advances in Neural Information Processing Systems}, volume~33,
  pages 1129--1141. Curran Associates, Inc., 2020.

\bibitem{zagoruyko2016wide}
Sergey Zagoruyko and Nikos Komodakis.
\newblock Wide residual networks.
\newblock In Edwin R.~Hancock Richard C.~Wilson and William A.~P. Smith,
  editors, {\em Proceedings of the British Machine Vision Conference (BMVC)},
  pages 87.1--87.12. BMVA Press, September 2016.

\bibitem{zeiler2012adadelta}
Matthew~D. Zeiler.
\newblock {ADADELTA:} an adaptive learning rate method.
\newblock {\em CoRR}, abs/1212.5701, 2012.

\bibitem{zhang2019theoretically}
Hongyang Zhang, Yaodong Yu, Jiantao Jiao, Eric Xing, Laurent~El Ghaoui, and
  Michael Jordan.
\newblock Theoretically principled trade-off between robustness and accuracy.
\newblock In Kamalika Chaudhuri and Ruslan Salakhutdinov, editors, {\em
  Proceedings of the 36th International Conference on Machine Learning},
  volume~97 of {\em Proceedings of Machine Learning Research}, pages
  7472--7482. PMLR, 09--15 Jun 2019.

\bibitem{zhang2019limitations}
Huan Zhang, Hongge Chen, Zhao Song, Duane~S. Boning, Inderjit~S. Dhillon, and
  Cho{-}Jui Hsieh.
\newblock The limitations of adversarial training and the blind-spot attack.
\newblock In {\em 7th International Conference on Learning Representations,
  {ICLR} 2019, New Orleans, LA, USA, May 6-9, 2019}. OpenReview.net, 2019.

\bibitem{zhang2018efficient}
Huan Zhang, Tsui-Wei Weng, Pin-Yu Chen, Cho-Jui Hsieh, and Luca Daniel.
\newblock Efficient neural network robustness certification with general
  activation functions.
\newblock In {\em Proceedings of the 32nd International Conference on Neural
  Information Processing Systems}, NIPS'18, page 4944–4953, Red Hook, NY,
  USA, 2018. Curran Associates Inc.

\bibitem{zhang2023proa}
Tianle Zhang, Wenjie Ruan, and Jonathan~E. Fieldsend.
\newblock Proa: A probabilistic robustness assessment against functional
  perturbations.
\newblock In Massih-Reza Amini, St{\'e}phane Canu, Asja Fischer, Tias Guns,
  Petra Kralj~Novak, and Grigorios Tsoumakas, editors, {\em Machine Learning
  and Knowledge Discovery in Databases}, pages 154--170, Cham, 2023. Springer
  Nature Switzerland.

\end{thebibliography}

\end{document}